\pdfoutput=1
\PassOptionsToPackage{table}{xcolor}
\documentclass[11pt]{article}

\usepackage[final]{acl}

\usepackage{times}
\usepackage{latexsym}

\usepackage[T1]{fontenc}

\usepackage[utf8]{inputenc}

\usepackage{microtype}

\usepackage{inconsolata}

\usepackage{hyperref}       
\usepackage{url}            
\usepackage{booktabs}       
\usepackage{amsfonts}       
\usepackage{nicefrac}       
\usepackage{microtype}      
\usepackage[table]{xcolor}
\usepackage{graphicx}
\usepackage{xspace}
\usepackage{amsmath}               

\usepackage{ragged2e}
\usepackage{makecell}
\usepackage{bbding}
\usepackage{wrapfig}
\usepackage{caption}
\usepackage{tcolorbox}
\usepackage{subcaption}
\usepackage{amssymb}
\usepackage{bbm}
\usepackage{enumitem}
\usepackage{multirow}
\usepackage{soul}
\usepackage{pifont}
\usepackage{xspace}

\definecolor{maroon}{cmyk}{0,0.87,0.68,0.32}
\definecolor{fullorange}{RGB}{236, 128, 60}
\definecolor{fullred}{RGB}{204, 0, 0}
\definecolor{fullgreen}{RGB}{106, 168, 79}
\definecolor{navy}{RGB}{0, 0, 128}
\definecolor{goldenrod}{RGB}{218, 165, 32}
\newcommand{\hlc}[2][yellow]{{%
    \colorlet{foo}{#1}%
    \sethlcolor{foo}\hl{#2}}%
}

\DeclareMathOperator*{\argmax}{argmax}

\newcommand{\model}{\textsc{SCORE}\xspace}

\title{Small Language Models Need Strong Verifiers to Self-Correct Reasoning}

\author{Yunxiang Zhang$^\alpha$\thanks{$\,\,\,$Correspondence to \tt{yunxiang@umich.edu}}\hspace{3pt}\quad \textbf{Muhammad Khalifa}$^\alpha$ \quad \textbf{Lajanugen Logeswaran}$^\beta$ \\
\textbf{Jaekyeom Kim}$^\beta$ \quad \textbf{Moontae Lee}$^{\beta\gamma}$  \quad
\textbf{Honglak Lee}$^{\alpha\beta}$ \quad \textbf{Lu Wang}$^\alpha$ \\
University of Michigan$^\alpha$ \quad LG AI Research$^\beta$ \quad University of Illinois at Chicago$^\gamma$
}

\begin{document}
\maketitle
\begin{abstract}

Self-correction has emerged as a promising solution to boost the reasoning performance of large language models (LLMs), where LLMs refine their solutions using self-generated critiques that pinpoint the errors.
This work explores whether small ($\leq 13$B) language models (LMs) have the ability of self-correction on reasoning tasks with minimal inputs from stronger LMs. We propose a novel pipeline that prompts smaller LMs to collect self-correction data that supports the training of self-refinement abilities. 
First, we leverage correct solutions to guide the model in critiquing their incorrect responses. 
Second, the generated critiques, after filtering, are used for supervised fine-tuning of the self-correcting reasoner through solution refinement. 
Our experimental results show improved self-correction abilities of two models on five datasets spanning math and commonsense reasoning, with notable performance gains when paired with a strong GPT-4-based verifier, though limitations are identified when using a weak self-verifier for determining when to correct.\footnote{Our implementation can be accessed at \url{https://github.com/yunx-z/SCORE}.}

\end{abstract}

\section{Introduction}
Recent research shows that large language models (LLMs)~\cite{DBLP:journals/corr/abs-2303-08774} can self-correct their responses to meet diverse user requirements, ranging from diminishing harmful content to including specific keywords and to debugging code~\cite{DBLP:journals/corr/abs-2303-17651,DBLP:journals/corr/abs-2304-05128}. 
Self-correction is typically accomplished by first generating a critique that identifies the shortcomings of the initial response, followed by revising it according to the self-critique---a process that can be iterated. 

Self-correction has emerged as an intriguing paradigm for rectifying the flaws in LLM's outputs~\cite{DBLP:journals/corr/abs-2308-03188}. However, models that are effective at self-correction are of very large sizes,
and many of them are proprietary and accessible only via APIs. 
In this work, we focus on the self-correction abilities of small, open-source language models (LMs).\footnote{While the distinction between small vs. large LMs is often context-dependent~\cite{DBLP:journals/corr/abs-2206-05802,DBLP:journals/corr/abs-2310-13522}, in this work, we interchangeably use ``small'' or ``weak'' LMs to refer to open models with a few billion parameters (e.g., LLaMA-7/13B~\cite{DBLP:journals/corr/abs-2307-09288}). 
}
Previous studies have shown that these smaller models can learn self-correction in reasoning through distillation from stronger LMs~\cite{DBLP:journals/corr/abs-2310-13522,DBLP:journals/corr/abs-2310-20689,DBLP:journals/corr/abs-2401-07301}. Yet this poses security risks for high-stakes domains and hinders the scientific understanding of enhancing LMs' ability to correct errors. 
We thus ask the question: 
\textit{To which degree do small LMs require guidance from strong LMs to learn self-correction for reasoning?}

We study this question by leveraging the small model itself to generate supervised training data to enhance its self-correction ability, instead of resorting to stronger LMs.
To this end, we draw inspiration from the rejection sampling fine-tuning (RFT)~\cite{DBLP:journals/corr/abs-2307-09288,DBLP:journals/corr/abs-2308-01825} method where LLM's reasoning skills are bootstrapped via diverse chain-of-thought sampling and supervised fine-tuning on the correct reasoning chains. 
We propose \textbf{\model}---an approach to bootstrap small LMs' \textbf{S}elf-\textbf{CO}rrection ability in \textbf{RE}asoning tasks. 
Concretely, we devise a pipeline for accumulating \textbf{high-quality critique-correction} data from small LMs, which are used for supervised fine-tuning of self-correcting reasoners.
First, we leverage correct solutions as hints for the base LMs to critique incorrect answers. By reverse-engineering from the correct answer, the models generate more effective critiques.
Second, we filter these critiques for correctness, well-formedness, and clarity using simple rule-based and prompting methods. 
Finally, we fine-tune the same LMs to become \textbf{self-refining} models using this curated data. 
By avoiding the use of supervision from stronger LMs, we ensure that our method enables a small LM to bootstrap its self-correction capabilities.

We evaluate our \model fine-tuned refiner under both extrinsic and intrinsic self-correction settings~\cite{DBLP:journals/corr/abs-2310-01798}.
The primary difference between these two settings is whether the refiner is allowed to use \textit{external} signals to determine \textit{when} to self-correct (i.e., refine the initial solution only when it is believed to be incorrect). 
Identifying \textit{when} to self-correct involves verifying the solutions' correctness, which is still challenging for current state-of-the-art LLMs without proper external feedback~\cite{DBLP:journals/corr/abs-2310-01798}.
We adopt a simple baseline for the self-verification problem following~\citet{DBLP:journals/corr/abs-2110-14168}. Specifically, we fine-tune the same LMs to become verifiers with labels based solely on the correctness of the final answer, conditioning on the question and a candidate solution.
As for the extrinsic setting, we simulate strong verifiers with GPT-4 and oracle labels, to show the effectiveness of small LMs as self-correcting reasoners.

We test the \model method with the \texttt{LLaMA-2-13B-chat}~\cite{DBLP:journals/corr/abs-2307-09288} and \texttt{Gemma-7b-it}~\cite{team2024gemma} models on five datasets spanning math and commonsense reasoning. 
We find that our model with \model fine-tuning
outperforms the original model by an average of 14.6\%
when using a \texttt{gpt-4}-based verifier. 
Nevertheless, the model struggles with self-correction when subjected to a weak self-verifier fine-tuned on self-generated solutions.

Our main contributions are summarized below:
\begin{enumerate}    
    \item We introduce \model, a novel pipeline to generate self-correction data from a small LM, and subsequently fine-tune the model to be a self-correcting reasoner.
    \item Our method effectively augments the self-correction abilities of small LMs on math and commonsense reasoning, when using strong verifiers.
    \item To the best of our knowledge, we are the first to demonstrate the potential of small LMs to bootstrap their abilities on self-corrective reasoning \textit{without} distilling training data from stronger LMs or using human annotation. 
\end{enumerate}


\begin{figure*}
\centering
\includegraphics[width=\linewidth]{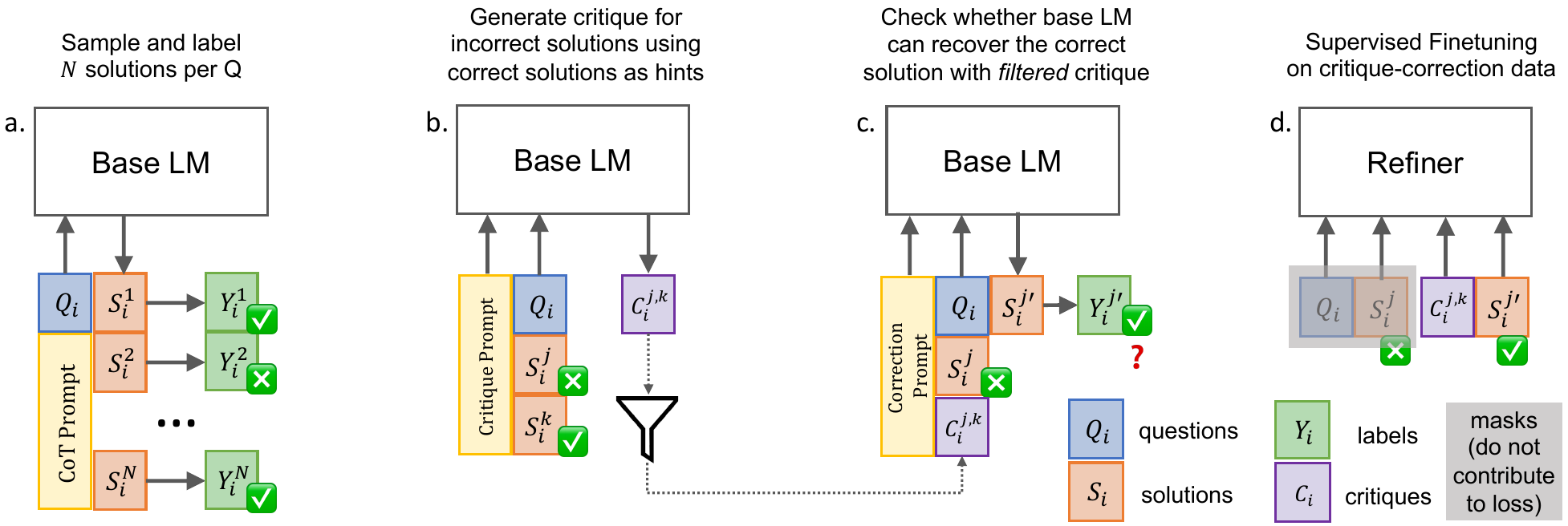}
\caption{
Illustration of the \model pipeline to generate critique-correction data from a small LM (step a-c) and fine-tune the same LM to self-correct its reasoning errors (step d), \textit{without} distilling any data from stronger LMs.}
\label{fig:method}
\end{figure*}

\section{Problem Formulation of Self-Correction}

\paragraph{Self-Correct := \textsc{(Self-)Verify} + \textsc{Self-Refine}.}

We decompose the task of self-correction into two phases: \textsc{(Self-)Verify} and \textsc{Self-Refine}. 
The LM first generates an initial solution for a reasoning question. 
A verifier, either the LM itself (intrinsic) or the external signal (extrinsic), then judges the correctness of the initial solution. 
If correct, the initial solution will be directly used as the final answer. If incorrect, a refiner will revise the solution. 
While this process can be iterated, we fix the times of iterations as 1 throughout this paper for efficiency and leave multiple iterations as future studies.

Decoupling \textsc{(Self-)Verify} and \textsc{Self-Refine} brings two major advantages over a one-model-does-all design. 
First, we can freely parameterize each module---for example, by using a fine-tuned and a few-shot prompted model. 
This allows us to carefully examine the impact of strong vs. weak verifiers on the refiners' performance. On the contrary, previous work on self-correction with small LMs~\cite{DBLP:journals/corr/abs-2310-13522,DBLP:journals/corr/abs-2310-20689,DBLP:journals/corr/abs-2401-07301} conflates \textsc{Self-Verify} and \textsc{Self-Refine}, creating a barrier to fully understanding the distinct capacities of these models in each skill.
Second, it reduces the difficulty of training each module, since the model only needs to specialize in one kind of ability, which is either verification or refinement.

\paragraph{\textsc{Self-Refine} := Critique + Correction.}
The challenge for \textsc{Self-Refine} is that it can be difficult for language models to directly map an initial solution to a revision without any guidance~\cite{DBLP:conf/iclr/WelleckLWBSK023}.
Using \textbf{critiques}---assessments that pinpoint the locations of errors within the reasoning steps, explain the causes of these errors, and offer guidance on how to correct them---can significantly enhance the performance of language models when generating revisions~\cite{DBLP:journals/corr/abs-2206-05802,DBLP:journals/corr/abs-2303-17651}.
Therefore, we formulate refinement with two steps: the model will first generate a critique for the initial solutions determined as incorrect, followed by a corrected version, in a single pass. 
Yet, it is still non-trivial to obtain high-quality critiques to guide the error correction. We address this problem using the correct solutions as hints to facilitate the critique generation, detailed in Section~\ref{sec:critique_generation}.

\section{The \model Method}
Our approach is inspired by rejection sampling fine-tuning (RFT): sampling diverse solutions for each question and fine-tune LLMs on the self-generated solutions that lead to the correct final answer~\cite{DBLP:journals/corr/abs-2308-01825,DBLP:conf/emnlp/0001GHW00023,DBLP:conf/nips/ZelikmanWMG22}. 
We want to bootstrap the small LM’s inherent ability to generate critiques for reasoning steps.
We design an end-to-end pipeline to collect self-correction data generated by small LMs at scale, without any distillation from stronger LMs. The self-generated critiques, after filtering, are used to fine-tune the smaller LM itself to bootstrap its ability to self-correct.
Concretely, the \model pipeline consists of two stages shown in Figure~\ref{fig:method} and described below. 

    \paragraph{Stage 1: Generating and Filtering Critiques.} 
    We sample $N$ solutions for each question in the training set by few-shot chain-of-thought prompting a base LM (step a). To enable the base LM to reflect on its incorrect solutions,
    we include a correct solution for the same question (if exists) in the prompt as a hint (step b). We then filter the self-generated critiques based on their correctness and clarity (step c). This process is detailed in Section~\ref{sec:critique_generation}.

    \paragraph{Stage 2: Supervised Fine-tuning of the Refiner.}
    The filtered critiques obtained from stage 1 are used in the next stage for fine-tuning the small LM itself.
    We train a refiner that generates critiques and corrections conditioned on questions and initial solutions (step d). We exclude the hints during fine-tuning. More details are given in Section~\ref{sec:verifier_refiner_ft}.

\subsection{Generating and Filtering Critiques}\label{sec:critique_generation}

Directly generating critique for an incorrect solution without external supervision signals is difficult. In our preliminary experiments, we find it easier for the LM to generate critiques \textbf{using correct solutions as hints},
as the model only needs to compare the different steps between these two solutions and justify the correct ones. In Appendix~\ref{app:proof}, we explain this intuition from a mathematical perspective.

To leverage correct solutions as hints for LMs to generate critiques on incorrect solutions, we label these solutions and collect all possible pairs of incorrect-correct solutions for the same questions (Cartesian product between the sets of incorrect and correct solutions). 
We craft a few-shot critique prompt (Appendix~\ref{sec:prompts}) to instruct the base LM to generate critiques for the incorrect solution using the paired correct solution as hints.\footnote{The total number of incorrect-correct solution pairs could be very large so we sample only one critique per pair. This has already provided a sufficient amount of \model fine-tuning data after filtering.}
Step-level critiques are more useful than solution-level ones since they provide more precise and fine-grained supervision ~\cite{DBLP:journals/corr/abs-2305-20050,DBLP:journals/corr/abs-2306-01693,uesato2022solving} that mitigate the undesirable behavior of LMs using incorrect reasoning to reach the correct final answer~\cite{khalifa-etal-2023-grace,DBLP:conf/nips/ZelikmanWMG22}.
Therefore, we prompt the model to provide feedback for each step of the initial solution, either endorsing the initial answer (e.g., ``\textit{this step is correct}'') or pinpointing the errors (e.g., ``\textit{there are errors in the step because ...}'').
To ensure the LM-generated critique is grounded in a specific step, we also ask the model to copy each step before providing feedback on it. 
Considering these requirements, we design the format of the critique prompt as follows, with a detailed example in Appendix~\ref{sec:prompts}. 

\begin{tcolorbox}[title=Critique Prompt]
\small
\texttt{Q: $\{\text{question}\}$}
\\

\texttt{Answer 1 (Incorrect):}

\texttt{Step 1: ...}

\texttt{...}

\texttt{Step $n$: The answer is \boxed{x}.}
\\

\texttt{Answer 2 (Correct):}

\texttt{Step 1: ...}

\texttt{...}

\texttt{Step $n$: The answer is \boxed{y}.}
\\

\texttt{There are reasoning errors in Answer 1. Please go through each
step in Answer 1, use Answer 2 as a reference for the correct
approach, and provide feedback that helps correct the errors in
Answer 1. End your response with [END].}
\\

\texttt{Let's go through the errors in Answer 1 and provide feedback:}
\\

\texttt{Answer 1 (Incorrect):}
\\

\texttt{Step 1: ...}

\texttt{Feedback: This step is correct.}

\texttt{...}

\texttt{Step $i$: ...}

\texttt{Feedback: This is incorrect. Because ...}

\texttt{...}

\texttt{Step $n$: The answer is \boxed{x}.}

\texttt{Feedback: The correct answer, based on the corrected calculations, should be \boxed{y}. [END]}
\end{tcolorbox}
Note that the model should suggest the corrected final answer, taken from the hint solution, as part of the feedback for the last step. This forces the model to explicitly leverage the information from the hint solution.

\paragraph{Filtering Generated Critiques.}
After obtaining the raw self-generated critiques, we want to remove the low-quality ones and keep the rest for fine-tuning LMs.
Thanks to the well-designed format of critiques, we can apply rule-based filters to remove generated critiques that do not follow the desired format. These criteria include:
\begin{itemize}
    \item The number of steps and feedbacks (counted by the appearances of ``\texttt{Step \{i\}:}'' and ``\texttt{Feedback:}'') should be the same.
    \item Each step should be exactly copied from the initial solution.
    \item The feedback for the last step should provide the correct answer.
\end{itemize}
The first two criteria check for the well-formedness of the critique and the third one focuses on the correctness aspect. A critique will be removed if it fails to meet any of the three criteria above. 

Given that a critique could still contain errors even if it suggests the correct final answer in the last step, we add an additional stage of prompting-based filtering besides the above rule-based heuristics. Specifically, we prompt the base LM to revise the incorrect solution given the critique that already passes the aforementioned filtering rule. 
Assuming the base LM has reasonable ability of following instructions, it is expected to give a correct revision if the generated critique is both clear and error-free. We demonstrate such an example of the correction prompt in Appendix~\ref{sec:prompts}. 
In other words, we remove critiques that do not result in a correctly revised answer. After the ruled-based and prompting-based filtering, we obtain the high-quality critiques
for fine-tuning LMs to self-refine.

\subsection{Supervised Fine-tuning of the Refiner}\label{sec:verifier_refiner_ft}

We train the refiner to generate a critique and an improved solution in one pass conditioned on a question and an initial solution. 
We note that although we provide the correct solutions as hints to generate critiques during data collection, the model is tasked to generate critiques \textit{without} the hints during fine-tuning and inference. 
Previously we collected critiques for every step to ensure that we can apply multiple filters to obtain high-quality critiques.
But in this step, we truncate the critiques to only keep the feedback for the first error step as the fine-tuning target.
This is because it is difficult to ask the LMs to identify and correct all the errors in one pass~\cite{DBLP:journals/corr/abs-2310-13522} without referring to correct solutions as hints during inference.
The refiner is fine-tuned on truncated critiques and corrections collected in the previous stages with cross-entropy loss.
We do not include few-shot demonstrations during fine-tuning. We apply masks on the input 
tokens so that they do not contribute to the loss. 
Although we only do 1 iteration for the refinement in this work, we later show that small LMs can already achieve great improvement after 1 round of self-correction when paired with a strong verifier.

\begin{table}[t]
\setlength\tabcolsep{3pt}\centering
\small
\begin{tabular}{lcccc}\toprule
&\multicolumn{2}{c}{GSM8K}  &\multicolumn{2}{c}{CSQA} \\\cmidrule{2-5}
& \# & \% & \# & \%\\\midrule
\multicolumn{3}{l}{\textit{Base LM: \texttt{LLaMA-2-13b-chat}}} \\\cmidrule{1-5}
Raw critiques &56,843& 100.0 &42,705&100.0 \\
After rule-based filering &36,337&63.9 &36,436&85.3 \\
\makecell[l]{After prompting filtering\\\quad (for \model fine-tuning)} &14,499&25.5 &24,511&57.4 \\\midrule
\multicolumn{3}{l}{\textit{Base LM: \texttt{Gemma-7b-it}}} \\\cmidrule{1-5}
Raw critiques &52,669&100.0 &40,604&100.0 \\
After rule-based filering &17,209&32.7 &35,929&88.5 \\
\makecell[l]{After prompting filtering\\\quad (for \model fine-tuning)} &4,623&8.8 &12,972&31.9 \\
\bottomrule
\end{tabular}
\caption{Statistics of the critique data generated from our \model pipeline. Although Gemma-7B has fewer data left after filtering, it still achieves greater improvement than LLaMA-13B by self-correction (Section~\ref{sec:main_findings}), suggesting that Gemma-7B is more effective at learning self-correction from \model.}
\label{tab:stat}
\end{table}

\section{Experimental Setup}

\paragraph{Self-Correction Data Collection.}
As stated in Section~\ref{sec:critique_generation}, we sample $N=10$ solutions from the base model with the chain-of-thought (CoT) prompts shown in Appendix~\ref{sec:prompts}, label their correctness, and formulate incorrect-correct solution pairs for critique generation. We separately collect data for each base LM and task. This results in the number of raw critiques shown in Table~\ref{tab:stat}.
We sequentially apply rule-based filtering and prompting-based filtering to obtain high-quality critiques for the fine-tuning data.

\paragraph{Verifiers.}
We experiment with verifiers of different levels of capabilities to gauge their impacts on self-correction performance.
First, we adopt a simple baseline for training a \textbf{self-verifier} following~\citet{DBLP:journals/corr/abs-2110-14168}. The self-verifier is a model with the same architecture as the base LM, conditioned on the question and a candidate solution to judge the probability that the solution is correct/incorrect. Specifically, we label the solutions sampled from the base LM as incorrect (0) or correct (1) solely based on their final answers and fine-tune the verifier with a binary classification head on the last-layer representation of the last token in the input sequence ``\texttt{Question: \{q\} \textbackslash n Solution: \{s\} \textbackslash n Is this solution correct?}''. Since the fine-tuning data is imbalanced between correct and incorrect solution, we re-weight the loss for each class with regarding to its proportion. 
During inference, the verifier model outputs a probability of the initial solution being incorrect, and the refinement is introduced only when the confidence of the verifier's predictions exceeds a certain threshold, which is automatically chosen in a way that maximizes the accuracy on the dev set and then fixed during test-time predictions. 
Since fine-tuned small LMs are still weak verifiers that bottleneck the performance of self-correction, we also experiment with a second option by using \textbf{\texttt{gpt-4}} as an off-the-shelf strong verifier to demonstrate the potential of our fine-tuned refiner. We do so by few-shot prompting \texttt{gpt-4} to predict the correctness of the initial solution by smaller LMs, with the verifying prompt shown in Appendix~\ref{sec:prompts}. Finally, we directly use the gold labels of the initial solutions as signals to determine when to self-refine. This \textbf{oracle} verifier setting provides an upper bound for the refiners' performance.

\paragraph{Benchmarks and Base Models.}
To demonstrate the effectiveness of the \model method, we conduct experiments on two popular datasets: \textbf{GSM8K}~\cite{DBLP:journals/corr/abs-2110-14168} for mathematical reasoning and \textbf{CommonsenseQA}~\cite{DBLP:journals/corr/abs-1811-00937} for commonsense reasoning. 
We also conduct transferability studies and evaluate the generalization performance of our fine-tuned refiner on \textbf{MATH}~\cite{DBLP:conf/nips/HendrycksBKABTS21} for mathematical reasoning, \textbf{QASC}~\cite{DBLP:conf/aaai/KhotCGJS20} and \textbf{RiddleSense}~\cite{DBLP:conf/acl/LinWYLR21} for commonsense reasoning.
Specifically, for mathematical reasoning, we train self-verifiers and \model refiners using only GSM8K training data and evaluate them on the whole GSM8K test set and a subset of MATH test set,\footnote{This subset includes a total 181 problems of Level 1 difficulty in MATH with question types of algebra, Counting \& probability, prealgebra and number theory, where the final answer is a number and no latex exists in the question.} following the practice of~\citet{Hosseini2024VSTaRTV}. 
Similarly, for commonsense reasoning, we fine-tune our models using only CommonsenseQA training data and evaluate them on the whole dev\footnote{The test labels of these datasets are hidden, so we use the original dev set as our test set, following~\citet{DBLP:conf/nips/KojimaGRMI22,DBLP:conf/nips/KimBM23}.} set of CommonsenseQA, QASC, and RiddleSense.
Since questions in CommonsenseQA, QASC, and Riddlesense have a multiple-choice format, we also include a random refiner baseline that randomly picks a choice different from the initial answer, following the practice of~\citet{DBLP:journals/corr/abs-2310-01798}.

\begin{figure*}
\centering
\includegraphics[width=\linewidth]{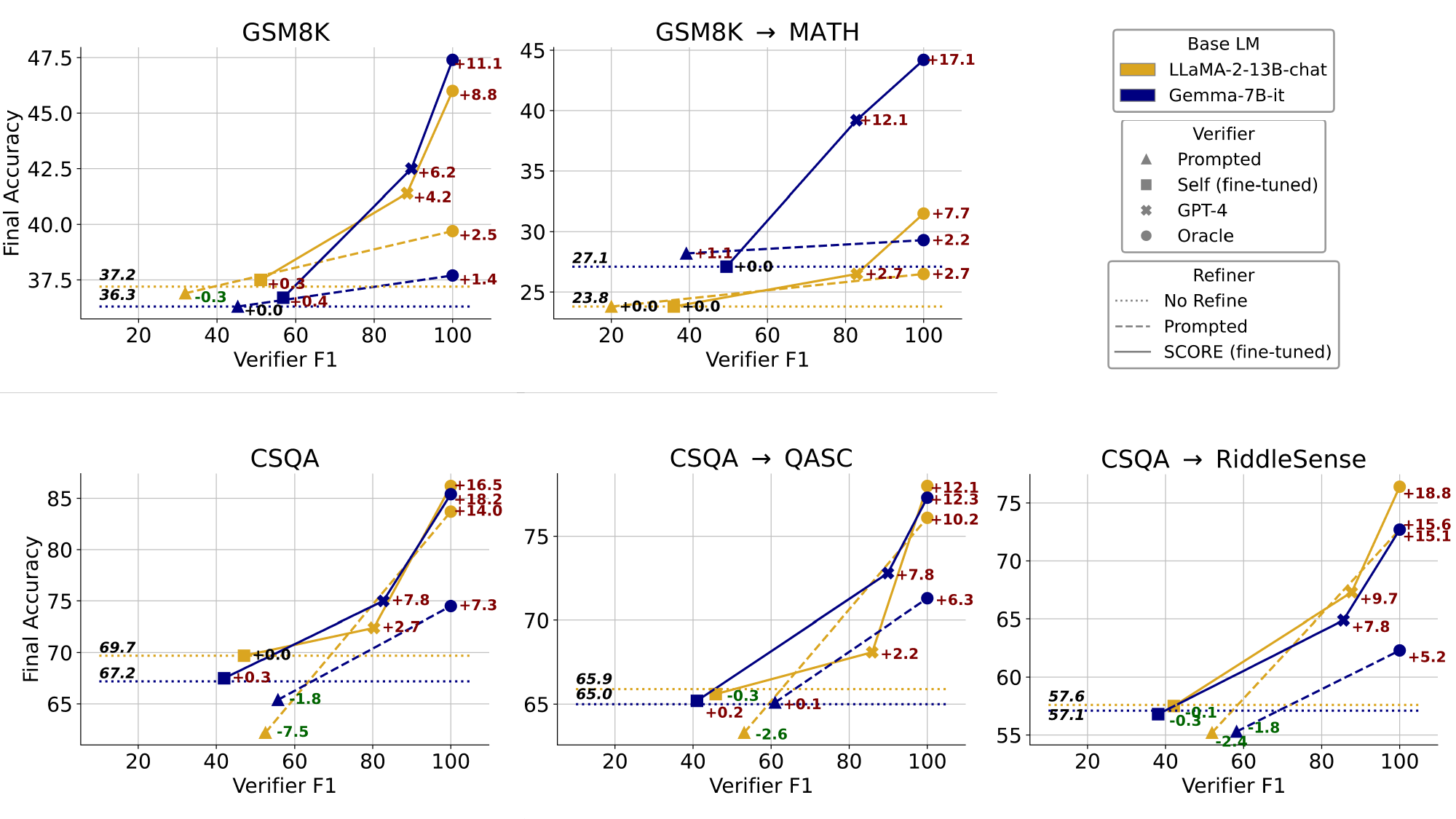}
\caption{
Performance of \model models using \textcolor{goldenrod}{\texttt{\textbf{LLaMA-2-13B-chat}}} and \textcolor{navy}{\texttt{\textbf{Gemma-7B-it}}} as base LMs.
We show the relationship between F1 scores of the verifiers and final answer accuracies. We also report the difference between the final accuracy after refinement and the corresponding initial answer accuracy (\textit{Init. Accu.}) by few-shot prompting for each base LM. For the concrete numbers of these metrics, please see Table~\ref{tab:new_main_results} in the appendix.
We include test results for training tasks (GSM8K and CommonsenseQA/CSQA), as well as transfer evaluation of GSM8K trained models on MATH subset, CSQA trained models on QASC and RiddleSense. 
All models use greedy decoding during inference. 
}
\label{fig:new_main_results}
\end{figure*}

We explore two open-source smaller language models, namely \texttt{LLaMA-2-13B-chat}~\cite{DBLP:journals/corr/abs-2307-09288} and \texttt{Gemma-7B-it}~\cite{team2024gemma} as the base LMs to generate self-correction data and evaluate their self-correction abilities. In Appendix~\ref{sec:rft_detail}, we also investigate whether our self-correction fine-tuning can be built on top of other fine-tuning methods (e.g., rejection-sampling fine-tuning) to further boost the reasoning performance.

\paragraph{Fine-tuning and Evaluation.}
We fine-tune the base LM using the LLaMA-Factory library~\cite{llama-factory} with LoRA~\cite{DBLP:conf/iclr/HuSWALWWC22}. We set the low-rank dimension as 32, the learning rate as 2e-5, training epochs as 3, batch size as 32. During inference, we set the temperature as 0 (i.e., greedy decoding) and the max sample length as 2,048. All our experiments can be conducted on 4$\times$A40 GPU with 48GB of memory.

\section{Results}

In this section, we first present the experimental findings of \model method on various models and datasets (Section~\ref{sec:main_findings}). To better understand the performance changes after self-correction, we then analyze the behaviors of verifiers and refiners (Section~\ref{sec:analysis_sc}) and further highlight several key design decisions of our pipeline with ablation studies (Section~\ref{sec:abalation}). Lastly, we show the impact of \model fine-tuning data size on self-correction performance (Section~\ref{sec:scaling}). 

\subsection{Main Findings}\label{sec:main_findings}
Figure~\ref{fig:new_main_results} presents the primary evaluation results for our fine-tuned models compared to baseline models. The results include two performance metrics: the verifier F1, which assesses the precision and recall of the verifier's predictions; and the final accuracy, which measures the accuracy of the final answer after self-correction.
We have four major findings.

\textit{1) The critique-correction data collected by our \model pipeline enhances the base LM's capability for self-correction.} Our fine-tuned models consistently bring large improvements on the final accuracy over the initial answer obtained by few-shot prompting. However, the prompting-based self-correction baseline (prompted verifier + prompted refiner in Figure~\ref{fig:new_main_results}) proposed  by~\citet{DBLP:journals/corr/abs-2303-17651} deteriorates the final predictions, as LMs struggle to identify errors in their reasoning~\cite{DBLP:journals/corr/abs-2310-01798} and possess limited self-correction abilities before bootstrapping. 
On the multiple-choice CommonsenseQA questions, our \model fine-tuned refiner achieves much larger improvement than the random baseline under oracle verifier, indicating that our model is not simply making random guess.

\begin{figure}[t]
\centering
\includegraphics[width=0.8\linewidth]{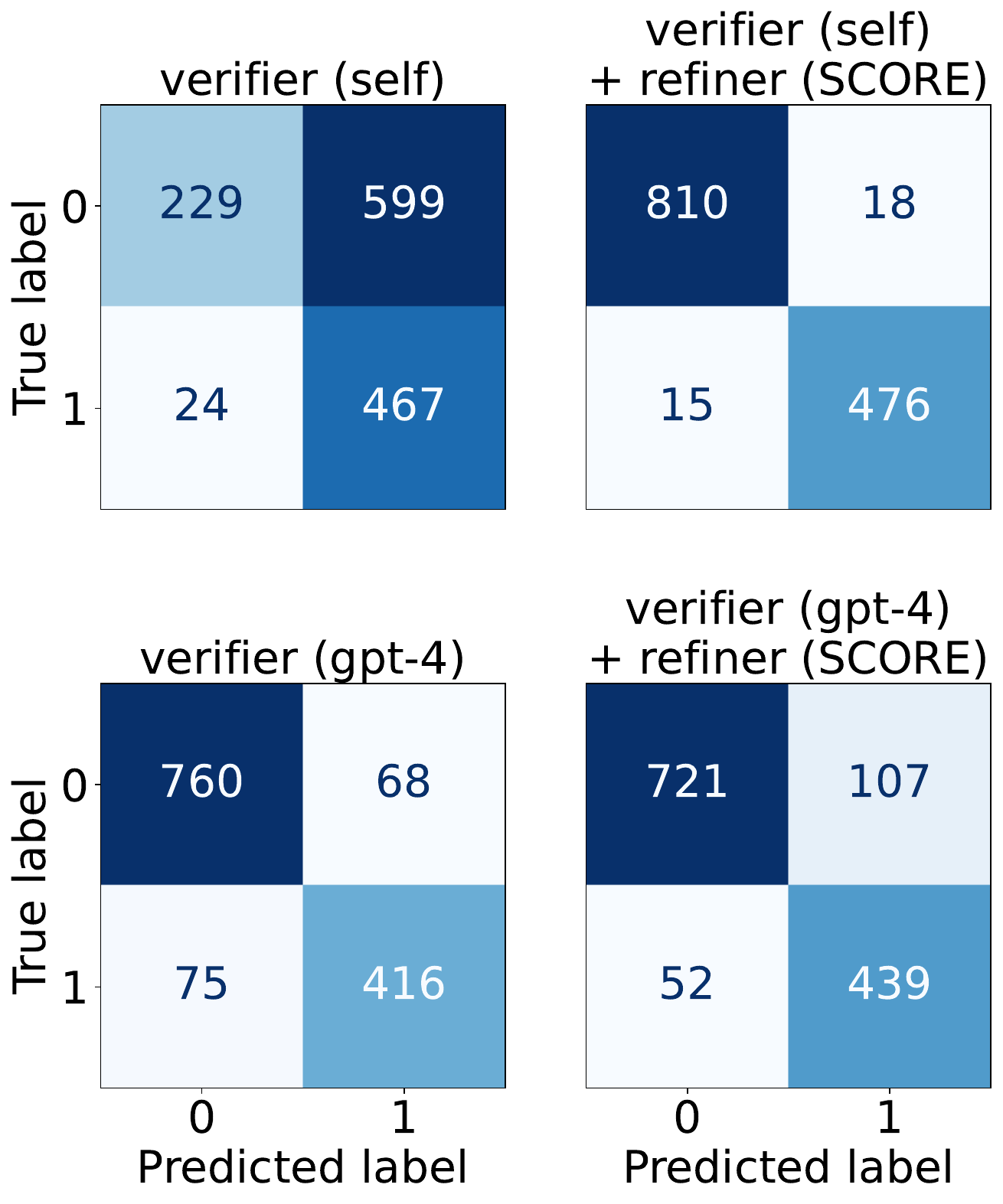}
\caption{
Confusion matrices of the predictions by the verifier and the refiner on GSM8K test set. The base LM is \texttt{LLaMA-2-13B-chat}. ``True label'' means the correctness of the initial solution. The predicted label of the verifier represents whether the verifier judges it as correct (1) or incorrect (0). The predicted label of the verifier + refiner is the correctness of the final answer. The strong verifier (\texttt{gpt-4}) makes fewer false positive predictions than the weak self-verifier and unleashes the potential of the small LM to revise an incorrect answer into a correct one more likely than the other way around.
}
\label{fig:confusion_matrix}
\end{figure}

\textit{2) Our framework improves self-correction for various base LMs on different types of reasoning tasks.} We validate the effectiveness of our \model fine-tuning on both math reasoning and commonsense reasoning tasks with two pretrained LMs. In principle, our task-agonistic pipeline can be applied to a variety of datasets whose reasoning could be expressed in a step-by-step format. We also observe that although the initial solutions proposed by Gemma-7B are worse than LLaMA-13B (e.g., 67.2 < 69.7 on CommonsenseQA), Gemma-7B's accuracy surpasses LLaMA-13B after self-correction (e.g., 75.0 > 72.4 on CommonsenseQA). Considering that Gemma-7B is fine-tuned with even less self-correction data (Table~\ref{tab:stat}), we believe Gemma is more effective at learning self-correction skills from \model fine-tuning.

\textit{3) The self-correction performance is largely bottlenecked by the verifier rather than the refiner.} 
Using the same fine-tuned refiner, the final accuracies vary a lot among different verifiers.
The upper bound performance suggested by an oracle verifier demonstrate great potential for self-correction, yet a weak self-verifier can only bring minor improvements, if not misguiding the refiner.
Nevertheless, when combined with a more advanced verifier, such as GPT-4, our refiner achieves a significant increase in final accuracy, e.g., an average of +8.3 across five datasets for \texttt{Gemma-7b-it}. 
The confusion matrices in Figure~\ref{fig:confusion_matrix} show the system of gpt-4-as-verifier + \model-refiner is more likely to modify an incorrect answer to a correct one than the other way round.
This observation underscores the necessity of effectively tackling the problem of reasoning verification before significant advances in self-correction can be attained. Future work could focus on the improvement of reasoning verification that is built upon a mechanistic~\cite{DBLP:journals/corr/abs-2309-15098,DBLP:journals/corr/abs-2402-16837} and representational~\cite{DBLP:journals/corr/abs-2310-01405,zheng2024prompt} understanding of LMs' internal reasoning process.

\textit{4) The enhanced self-correction skills can transfer across different datasets.} When evaluating our fine-tuned refiner on unseen datasets, it still demonstrates consistent improvement over the baselines (up to +12.1 by the GSM8K-trained Gemma-7B on MATH subset). This shows that the model is learning generalizable self-correction skills rather than overfitting to a specific dataset. Additionally, we find that the verifier does not transfer as well as the refiner, reiterating the difficulty of reasoning verification for LMs.

\model could be combined with oversample-then-rerank, an orthogonal approach to further improve reasoning~\cite{DBLP:journals/corr/abs-2110-14168,DBLP:conf/acl/LiLZFCLC23,Hosseini2024VSTaRTV}.
By aggregating verifications from multiple solutions, the weak verifiers become stronger via ensembling and thus unleash the power of refiner for better self-correction of reasoning. Please refer to Appendix~\ref{sec:comp_rerank} for details.

\subsection{Analysis of Self-Correction Behaviors}\label{sec:analysis_sc}
Following the methodology of \citet{DBLP:journals/corr/abs-2310-13522}, we focus on two key metrics to understand the model's self-correction behaviors: 1) the frequency with which the verifier decides to self-correct (Freq.), and 2) the extent to which these self-correction attempts enhance the model's task performance (Contrib.). Self-correction Freq. is measured by the ratio of self-correction attempts to the size of the test set, while self-correction Contrib. is determined by the number of instances in which these attempts successfully resulted in the correct answer.

Table~\ref{tab:more_results} presents a detailed analysis of the model's self-correction behaviors. Our analysis demonstrates that our fine-tuned refiner has a higher contribution to the final self-correction performance, explaining why it outperforms the prompting-based refiner~\cite{DBLP:journals/corr/abs-2303-17651}, as shown in Figure~\ref{fig:new_main_results}. 
Additionally, we find that
our fine-tuned verifier and the \texttt{gpt-4} verifier maintain a more reasonable frequency of self-correction, striking a better balance between correction attempts and accuracy.

\begin{table}[t]
\setlength\tabcolsep{3pt}\centering
\small
\begin{tabular}{lc|cccc}\toprule
\multirow{2}{*}{Verifier} &\multirow{2}{*}{Refiner} &\multicolumn{2}{c}{GSM8K} &\multicolumn{2}{c}{CSQA} \\\cmidrule{3-6}
 & &Freq. &Contrib. &Freq. &Contrib. \\\midrule
  \multicolumn{4}{l}{\textit{Base LM: \texttt{LLaMA-2-13b-chat}}} \\\cmidrule{1-6}
prompted &\multirow{3}{*}{prompted} &3.7 &10.2 & 17.5 & 19.6 \\
self & &2.7 &2.9 & 1.2 & 40.0  \\
oracle & &62.8 &4.0 & 30.3 & 46.2 \\\cmidrule{1-2}
self &\multirow{3}{*}{\makecell[c]{\model\\(fine-tuned)}} &19.0 &10.8 & 3.0 & 33.3 \\
\texttt{gpt-4} & & 63.3 & 15.6 & 38.8 & 40.9 \\
oracle & &62.8 &14.0 & 30.3 & 54.3 \\\midrule
  \multicolumn{4}{l}{\textit{Base LM: \texttt{Gemma-7b-it}}} \\\cmidrule{1-6}
prompted &\multirow{3}{*}{prompted} &18.7 &21.5 & 20.2 & 45.3 \\
self & &9.9 &9.9 & 2.9 & 42.9 \\
oracle & &63.7 &2.1 & 32.8 & 22.4\\\cmidrule{1-2}
self &\multirow{3}{*}{\makecell[c]{\model\\(fine-tuned)}} &27.9 &14.1 & 0.6 & 57.1 \\
\texttt{gpt-4} & & 63.4 & 17.1 & 38.6 & 48.4 \\
oracle & &63.7 &17.4 & 32.8 & 55.6 \\
\bottomrule
\end{tabular}
\caption{
Analysis of self-correction behaviors. The settings are the same as those in Figure~\ref{fig:new_main_results}.  Freq. (in percentage) means the frequency with which the verifier decides to self-correct. Contrib. (in percentage) refers to the extent to which these self-correction attempts enhance the model's task performance.
A strong verifier (e.g., \texttt{gpt-4}) maintains a balanced frequency of self-correction, ideally similar to that of the oracle verifier, and thus enhance the final accuracies in Figure~\ref{fig:new_main_results}.
Additionally, the SCORE refiner possesses stronger refinement capabilities, as indicated by a higher contribution score than the prompted refiner.
}\label{tab:more_results}
\end{table}

\begin{table}\centering
\small
\begin{tabular}{cccc}\toprule
\makecell[c]{Critique for\\\textit{only} the first\\error step?} &\makecell[c]{Generating\\ critique \&\\correction\\in \textit{one} pass?} &\makecell[c]{Final\\Accu.} &\makecell[c]{Oracle\\Accu.} \\\cmidrule{1-4}
\multicolumn{2}{c}{\makecell[c]{Initial answers by\\few-shot prompting}} &39.4 &39.4 \\\cmidrule{1-4}
\Checkmark &\Checkmark &\textbf{40.2} &\textbf{49.5} \\
\XSolidBrush &\Checkmark &39.6 &46.4 \\
\Checkmark &\XSolidBrush &39.4 &43.6 \\
\XSolidBrush &\XSolidBrush &39.9 &47.0 \\
\bottomrule
\end{tabular}
\caption{
Ablation study on the format of critique and decoupling critique and correction generation. Results are shown on GSM8K dev set with \texttt{LLaMA-2-13B-chat} as base LM.
}\label{tab:ablation_refiner}
\end{table}

\subsection{Ablation Studies}\label{sec:abalation}
In order to validate the various design decisions made in constructing our pipeline, we have conducted a series of ablation studies. The key findings from Table~\ref{tab:ablation_refiner} can be summarized as follows.  
1) It is easier for the LM to identify only the first erroneous step, as the performance drops if we challenge it to critique every step.
2) There is no need to separate the \textsc{Self-Refine} process into two modules---one for generating critiques and another for corrections. Such a separation not only increases the system's complexity and delays inference but also leads to a diminished final accuracy.

\subsection{Scaling with Fine-tuning Data Size}\label{sec:scaling}

\begin{figure}
\centering
\includegraphics[width=\linewidth]{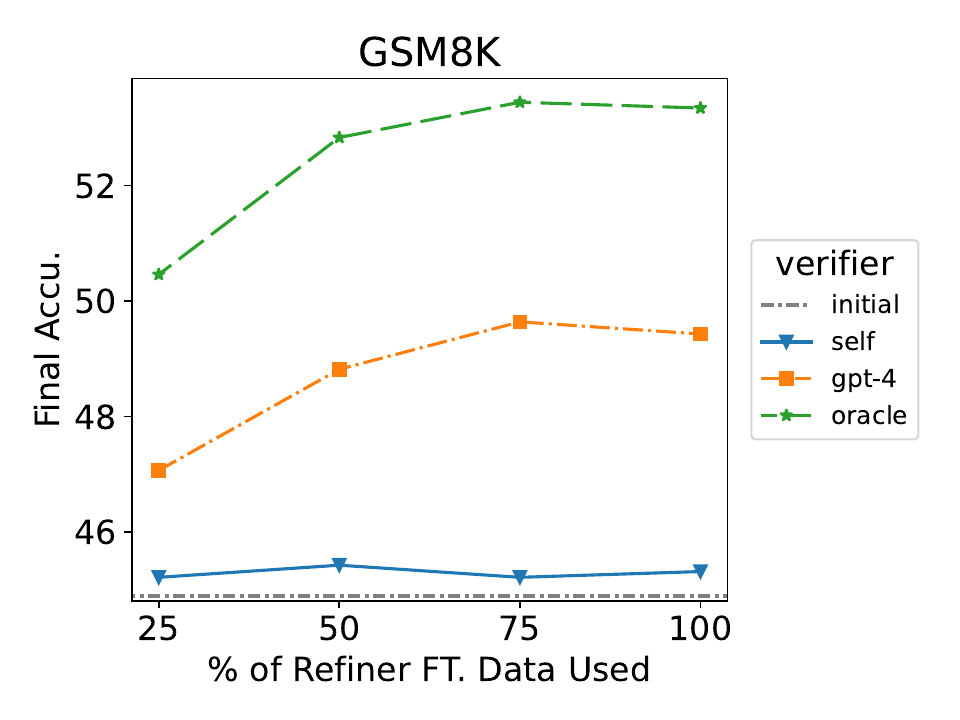}
\caption{Final accuracy on GSM8K dev set w.r.t percentage of refiner fine-tuning data used and the type of verifier. The base LM is \texttt{LLaMA-2-13B-chat}. Under a strong verifier setting, our refiner brings greater performance gain with more fine-tuning data.
Yet the performance plateaus with a weak verifier.
}
\label{fig:gsm8k_scale}
\end{figure}

We investigate the data-efficiency for refiner fine-tuning. Figure~\ref{fig:gsm8k_scale} plots the size of the refiner fine-tuning data against the final accuracy with different verification settings on GSM8K. We fine-tune the \texttt{LLaMA-2-13B-chat} base model on a random subset (varying from 25\% to 75\%) of the 14,499 total critique-corrections as previously shown in Table~\ref{tab:stat}. We find that our refiner benefits from more fine-tuning data when paired with strong verifiers (oracle labels or \texttt{gpt-4}). Yet this effect is not observed when using a weak self-verifier, again highlighting the importance of verification for self-correction. 
We find that increasing the fine-tuning dataset size yields accuracy improvements up to a certain point. For instance, beyond approximately 10k examples (representing 75\% of the \model fine-tuning data), the performance does not further improve.

\section{Related Work}

\paragraph{Training Small Language Model to Self-Correct.}
Recent work shows that smaller language model can be fine-tuned on task-specific data to perform self-correct. But existing methods rely either on distilled data from stronger models~\cite{DBLP:journals/corr/abs-2310-20689,DBLP:journals/corr/abs-2310-13522,DBLP:journals/corr/abs-2401-07301,selfee2023,Zhang2024ULTRAUL} or template-based critiques~\cite{DBLP:journals/corr/abs-2304-01904,DBLP:conf/iclr/WelleckLWBSK023}.  Our approach differs from prior studies in this domain as we gather natural language critiques from a small language model without relying on larger models or task-specific heuristics. Furthermore, we split the self-correction process into two phases: \textsc{(Self-)Verify} and \textsc{Self-Refine}. This separation contrasts with earlier approaches that often merge the two skills, which not only obscures the true abilities of these models in each respective skill but also complicates the training process. In a nutshell, we demonstrate that strong verifiers unleash the power of small LMs to \textsc{Self-Refine}.

\paragraph{Bootstrapping Reasoning in Language Models.}
As language models become more powerful, human supervision may not be sufficient to improve these models. This trend calls for self-improving LMs that provide training signals for themselves~\cite{DBLP:conf/nips/ZelikmanWMG22,DBLP:journals/corr/abs-2308-08998,DBLP:journals/corr/abs-2308-01825,DBLP:journals/corr/abs-2309-02144,DBLP:journals/corr/abs-2401-01335}. The bootstrapping methods often involve iteratively fine-tuning a base LM on its self-generated examples that obtain a high reward value for correctness, helpfulness, or other desired properties. The bootstrapping process can further leverage label-free data~\cite{DBLP:conf/emnlp/0001GHW00023,DBLP:journals/corr/abs-2308-06259,DBLP:journals/corr/abs-2401-10020} by generating pseudo labels using LLMs themselves. We draw inspiration from this family of methods and bootstrap the self-correction ability of smaller LMs.
Our method is complementary to the rejection-sampling finetuning approach and can further improve reasoning performance upon that.

\paragraph{Verifying Reasoning.}
Verification of reasoning chains involves judging the correctness of the final answer and each reasoning steps.
The verifier is often used to rerank multiple over-generated solutions and select the best one as the final output~\cite{DBLP:journals/corr/abs-2110-14168}, or guide the LLM decoding through the search space for correct reasoning paths~\cite{khalifa-etal-2023-grace}. We leverage a verifier to determine when to self-correct. Verifiers come at different granularity, including process/step-based ~\cite{DBLP:journals/corr/abs-2211-14275,DBLP:conf/acl/LiLZFCLC23,DBLP:journals/corr/abs-2305-20050} and outcome-based supervision~\cite{DBLP:journals/corr/abs-2110-14168,DBLP:journals/corr/abs-2311-09724,Hosseini2024VSTaRTV}. We use the latter since it is easier to construct labels automatically. 
Besides training a verifier with supervision signals, LLMs can also be few-shot prompted to become a verifier~\cite{DBLP:conf/emnlp/WengZX0HLSLZ23,DBLP:journals/corr/abs-2303-17651,DBLP:journals/corr/abs-2308-07921,DBLP:journals/corr/abs-2310-11511}. We also explores the possibility of LLM-as-verifier, and demonstrate its usage for self-correction.
We highlight the importance of verification in the context of self-correction, which echoes the recent finding that LLMs can successfully solve a problem, but cannot verify the reasoning~\cite{DBLP:journals/corr/abs-2402-19475,DBLP:journals/corr/abs-2402-06204,DBLP:journals/corr/abs-2311-00059}. This calls for more effort for building evaluation benchmarks ~\cite{DBLP:journals/corr/abs-2402-00559,DBLP:conf/nips/ChenZZCG0H23,DBLP:journals/corr/abs-2401-06961,DBLP:journals/corr/abs-2305-20050} and developing methods~\cite{DBLP:journals/corr/abs-2402-11199,DBLP:journals/corr/abs-2402-10528,Hosseini2024VSTaRTV} to improve the reasoning verification.

\section{Conclusion}
In this study, we investigate how to leverage minimal signals from strong LMs to teach small LMs to self-correct their reasoning. We propose the \model method to collect self-correction fine-tuning data solely from small LMs. We find that \model-fine-tuned small LMs become better refiner models without relying on knowledge distillation from stronger LMs, yet they still need strong verifiers to be successful at self-correcting their reasoning errors. 
Our results highlight that the self-verification limitation of LMs currently poses an obstacle to the advancement of intrinsic self-correction in reasoning and thus warrants future research.

\clearpage
\section*{Limitations}

Generating large amounts of synthetic data from smaller LMs requires intensive GPU computations, yet it removes the reliance on proprietary API models. Comparing the cost-efficiency of these two approaches will help us better trade-off between data generated from smaller LMs and larger LMs. Introducing the verifier and refiner during inference also causes additional latency, which we discuss in Appendix~\ref{sec:latency}.

\section*{Acknowledgements}
This work is supported by LG AI Research. We also thank ARR reviewers for their useful feedback.

\bibliography{custom}

\begin{thebibliography}{55}
\expandafter\ifx\csname natexlab\endcsname\relax\def\natexlab#1{#1}\fi

\bibitem[{An et~al.(2023)An, Ma, Lin, Zheng, Lou, and Chen}]{DBLP:journals/corr/abs-2310-20689}
Shengnan An, Zexiong Ma, Zeqi Lin, Nanning Zheng, Jian{-}Guang Lou, and Weizhu Chen. 2023.
\newblock \href {https://doi.org/10.48550/ARXIV.2310.20689} {Learning from mistakes makes {LLM} better reasoner}.
\newblock \emph{CoRR}, abs/2310.20689.

\bibitem[{Asai et~al.(2023)Asai, Wu, Wang, Sil, and Hajishirzi}]{DBLP:journals/corr/abs-2310-11511}
Akari Asai, Zeqiu Wu, Yizhong Wang, Avirup Sil, and Hannaneh Hajishirzi. 2023.
\newblock \href {https://doi.org/10.48550/ARXIV.2310.11511} {Self-rag: Learning to retrieve, generate, and critique through self-reflection}.
\newblock \emph{CoRR}, abs/2310.11511.

\bibitem[{Chen et~al.(2023{\natexlab{a}})Chen, Zhao, Zhang, Chern, Gao, Liu, and He}]{DBLP:conf/nips/ChenZZCG0H23}
Shiqi Chen, Yiran Zhao, Jinghan Zhang, I{-}Chun Chern, Siyang Gao, Pengfei Liu, and Junxian He. 2023{\natexlab{a}}.
\newblock \href {http://papers.nips.cc/paper\_files/paper/2023/hash/8b8a7960d343e023a6a0afe37eee6022-Abstract-Datasets\_and\_Benchmarks.html} {{FELM:} benchmarking factuality evaluation of large language models}.
\newblock In \emph{Advances in Neural Information Processing Systems 36: Annual Conference on Neural Information Processing Systems 2023, NeurIPS 2023, New Orleans, LA, USA, December 10 - 16, 2023}.

\bibitem[{Chen et~al.(2023{\natexlab{b}})Chen, Lin, Sch{\"{a}}rli, and Zhou}]{DBLP:journals/corr/abs-2304-05128}
Xinyun Chen, Maxwell Lin, Nathanael Sch{\"{a}}rli, and Denny Zhou. 2023{\natexlab{b}}.
\newblock \href {https://doi.org/10.48550/ARXIV.2304.05128} {Teaching large language models to self-debug}.
\newblock \emph{CoRR}, abs/2304.05128.

\bibitem[{Chen et~al.(2024)Chen, Deng, Yuan, Ji, and Gu}]{DBLP:journals/corr/abs-2401-01335}
Zixiang Chen, Yihe Deng, Huizhuo Yuan, Kaixuan Ji, and Quanquan Gu. 2024.
\newblock \href {https://doi.org/10.48550/ARXIV.2401.01335} {Self-play fine-tuning converts weak language models to strong language models}.
\newblock \emph{CoRR}, abs/2401.01335.

\bibitem[{Cobbe et~al.(2021)Cobbe, Kosaraju, Bavarian, Chen, Jun, Kaiser, Plappert, Tworek, Hilton, Nakano, Hesse, and Schulman}]{DBLP:journals/corr/abs-2110-14168}
Karl Cobbe, Vineet Kosaraju, Mohammad Bavarian, Mark Chen, Heewoo Jun, Lukasz Kaiser, Matthias Plappert, Jerry Tworek, Jacob Hilton, Reiichiro Nakano, Christopher Hesse, and John Schulman. 2021.
\newblock \href {http://arxiv.org/abs/2110.14168} {Training verifiers to solve math word problems}.
\newblock \emph{CoRR}, abs/2110.14168.

\bibitem[{Gu et~al.(2024)Gu, Li, Jain, Olausson, Lee, Sen, and Solar{-}Lezama}]{DBLP:journals/corr/abs-2402-19475}
Alex Gu, Wen{-}Ding Li, Naman Jain, Theo~X. Olausson, Celine Lee, Koushik Sen, and Armando Solar{-}Lezama. 2024.
\newblock \href {https://doi.org/10.48550/ARXIV.2402.19475} {The counterfeit conundrum: Can code language models grasp the nuances of their incorrect generations?}
\newblock \emph{CoRR}, abs/2402.19475.

\bibitem[{G{\"{u}}l{\c{c}}ehre et~al.(2023)G{\"{u}}l{\c{c}}ehre, Paine, Srinivasan, Konyushkova, Weerts, Sharma, Siddhant, Ahern, Wang, Gu, Macherey, Doucet, Firat, and de~Freitas}]{DBLP:journals/corr/abs-2308-08998}
{\c{C}}aglar G{\"{u}}l{\c{c}}ehre, Tom~Le Paine, Srivatsan Srinivasan, Ksenia Konyushkova, Lotte Weerts, Abhishek Sharma, Aditya Siddhant, Alex Ahern, Miaosen Wang, Chenjie Gu, Wolfgang Macherey, Arnaud Doucet, Orhan Firat, and Nando de~Freitas. 2023.
\newblock \href {https://doi.org/10.48550/ARXIV.2308.08998} {Reinforced self-training (rest) for language modeling}.
\newblock \emph{CoRR}, abs/2308.08998.

\bibitem[{Han et~al.(2024)Han, Liang, Shi, He, and Xiao}]{DBLP:journals/corr/abs-2401-07301}
Haixia Han, Jiaqing Liang, Jie Shi, Qianyu He, and Yanghua Xiao. 2024.
\newblock \href {https://doi.org/10.48550/ARXIV.2401.07301} {Small language model can self-correct}.
\newblock \emph{CoRR}, abs/2401.07301.

\bibitem[{Hendrycks et~al.(2021)Hendrycks, Burns, Kadavath, Arora, Basart, Tang, Song, and Steinhardt}]{DBLP:conf/nips/HendrycksBKABTS21}
Dan Hendrycks, Collin Burns, Saurav Kadavath, Akul Arora, Steven Basart, Eric Tang, Dawn Song, and Jacob Steinhardt. 2021.
\newblock \href {https://datasets-benchmarks-proceedings.neurips.cc/paper/2021/hash/be83ab3ecd0db773eb2dc1b0a17836a1-Abstract-round2.html} {Measuring mathematical problem solving with the {MATH} dataset}.
\newblock In \emph{Proceedings of the Neural Information Processing Systems Track on Datasets and Benchmarks 1, NeurIPS Datasets and Benchmarks 2021, December 2021, virtual}.

\bibitem[{Hosseini et~al.(2024)Hosseini, Yuan, Malkin, Courville, Sordoni, and Agarwal}]{Hosseini2024VSTaRTV}
Arian Hosseini, Xingdi Yuan, Nikolay Malkin, Aaron~C. Courville, Alessandro Sordoni, and Rishabh Agarwal. 2024.
\newblock \href {https://doi.org/10.48550/ARXIV.2402.06457} {V-star: Training verifiers for self-taught reasoners}.
\newblock \emph{CoRR}, abs/2402.06457.

\bibitem[{Hu et~al.(2022)Hu, Shen, Wallis, Allen{-}Zhu, Li, Wang, Wang, and Chen}]{DBLP:conf/iclr/HuSWALWWC22}
Edward~J. Hu, Yelong Shen, Phillip Wallis, Zeyuan Allen{-}Zhu, Yuanzhi Li, Shean Wang, Lu~Wang, and Weizhu Chen. 2022.
\newblock \href {https://openreview.net/forum?id=nZeVKeeFYf9} {Lora: Low-rank adaptation of large language models}.
\newblock In \emph{The Tenth International Conference on Learning Representations, {ICLR} 2022, Virtual Event, April 25-29, 2022}. OpenReview.net.

\bibitem[{Huang et~al.(2023{\natexlab{a}})Huang, Gu, Hou, Wu, Wang, Yu, and Han}]{DBLP:conf/emnlp/0001GHW00023}
Jiaxin Huang, Shixiang Gu, Le~Hou, Yuexin Wu, Xuezhi Wang, Hongkun Yu, and Jiawei Han. 2023{\natexlab{a}}.
\newblock \href {https://aclanthology.org/2023.emnlp-main.67} {Large language models can self-improve}.
\newblock In \emph{Proceedings of the 2023 Conference on Empirical Methods in Natural Language Processing, {EMNLP} 2023, Singapore, December 6-10, 2023}, pages 1051--1068. Association for Computational Linguistics.

\bibitem[{Huang et~al.(2023{\natexlab{b}})Huang, Chen, Mishra, Zheng, Yu, Song, and Zhou}]{DBLP:journals/corr/abs-2310-01798}
Jie Huang, Xinyun Chen, Swaroop Mishra, Huaixiu~Steven Zheng, Adams~Wei Yu, Xinying Song, and Denny Zhou. 2023{\natexlab{b}}.
\newblock \href {https://doi.org/10.48550/ARXIV.2310.01798} {Large language models cannot self-correct reasoning yet}.
\newblock \emph{CoRR}, abs/2310.01798.

\bibitem[{Jacovi et~al.(2024)Jacovi, Bitton, Bohnet, Herzig, Honovich, Tseng, Collins, Aharoni, and Geva}]{DBLP:journals/corr/abs-2402-00559}
Alon Jacovi, Yonatan Bitton, Bernd Bohnet, Jonathan Herzig, Or~Honovich, Michael Tseng, Michael Collins, Roee Aharoni, and Mor Geva. 2024.
\newblock \href {https://doi.org/10.48550/ARXIV.2402.00559} {A chain-of-thought is as strong as its weakest link: {A} benchmark for verifiers of reasoning chains}.
\newblock \emph{CoRR}, abs/2402.00559.

\bibitem[{Khalifa et~al.(2023)Khalifa, Logeswaran, Lee, Lee, and Wang}]{khalifa-etal-2023-grace}
Muhammad Khalifa, Lajanugen Logeswaran, Moontae Lee, Honglak Lee, and Lu~Wang. 2023.
\newblock \href {https://doi.org/10.18653/v1/2023.findings-emnlp.1022} {{GRACE}: Discriminator-guided chain-of-thought reasoning}.
\newblock In \emph{Findings of the Association for Computational Linguistics: EMNLP 2023}, pages 15299--15328, Singapore. Association for Computational Linguistics.

\bibitem[{Khot et~al.(2020)Khot, Clark, Guerquin, Jansen, and Sabharwal}]{DBLP:conf/aaai/KhotCGJS20}
Tushar Khot, Peter Clark, Michal Guerquin, Peter Jansen, and Ashish Sabharwal. 2020.
\newblock \href {https://doi.org/10.1609/AAAI.V34I05.6319} {{QASC:} {A} dataset for question answering via sentence composition}.
\newblock In \emph{The Thirty-Fourth {AAAI} Conference on Artificial Intelligence, {AAAI} 2020, The Thirty-Second Innovative Applications of Artificial Intelligence Conference, {IAAI} 2020, The Tenth {AAAI} Symposium on Educational Advances in Artificial Intelligence, {EAAI} 2020, New York, NY, USA, February 7-12, 2020}, pages 8082--8090. {AAAI} Press.

\bibitem[{Kim et~al.(2023)Kim, Baldi, and McAleer}]{DBLP:conf/nips/KimBM23}
Geunwoo Kim, Pierre Baldi, and Stephen McAleer. 2023.
\newblock \href {http://papers.nips.cc/paper\_files/paper/2023/hash/7cc1005ec73cfbaac9fa21192b622507-Abstract-Conference.html} {Language models can solve computer tasks}.
\newblock In \emph{Advances in Neural Information Processing Systems 36: Annual Conference on Neural Information Processing Systems 2023, NeurIPS 2023, New Orleans, LA, USA, December 10 - 16, 2023}.

\bibitem[{Kojima et~al.(2022)Kojima, Gu, Reid, Matsuo, and Iwasawa}]{DBLP:conf/nips/KojimaGRMI22}
Takeshi Kojima, Shixiang~Shane Gu, Machel Reid, Yutaka Matsuo, and Yusuke Iwasawa. 2022.
\newblock \href {http://papers.nips.cc/paper\_files/paper/2022/hash/8bb0d291acd4acf06ef112099c16f326-Abstract-Conference.html} {Large language models are zero-shot reasoners}.
\newblock In \emph{Advances in Neural Information Processing Systems 35: Annual Conference on Neural Information Processing Systems 2022, NeurIPS 2022, New Orleans, LA, USA, November 28 - December 9, 2022}.

\bibitem[{Li et~al.(2023{\natexlab{a}})Li, Yu, Zhou, Schick, Zettlemoyer, Levy, Weston, and Lewis}]{DBLP:journals/corr/abs-2308-06259}
Xian Li, Ping Yu, Chunting Zhou, Timo Schick, Luke Zettlemoyer, Omer Levy, Jason Weston, and Mike Lewis. 2023{\natexlab{a}}.
\newblock \href {https://doi.org/10.48550/ARXIV.2308.06259} {Self-alignment with instruction backtranslation}.
\newblock \emph{CoRR}, abs/2308.06259.

\bibitem[{Li et~al.(2023{\natexlab{b}})Li, Lin, Zhang, Fu, Chen, Lou, and Chen}]{DBLP:conf/acl/LiLZFCLC23}
Yifei Li, Zeqi Lin, Shizhuo Zhang, Qiang Fu, Bei Chen, Jian{-}Guang Lou, and Weizhu Chen. 2023{\natexlab{b}}.
\newblock \href {https://doi.org/10.18653/V1/2023.ACL-LONG.291} {Making language models better reasoners with step-aware verifier}.
\newblock In \emph{Proceedings of the 61st Annual Meeting of the Association for Computational Linguistics (Volume 1: Long Papers), {ACL} 2023, Toronto, Canada, July 9-14, 2023}, pages 5315--5333. Association for Computational Linguistics.

\bibitem[{Lightman et~al.(2023)Lightman, Kosaraju, Burda, Edwards, Baker, Lee, Leike, Schulman, Sutskever, and Cobbe}]{DBLP:journals/corr/abs-2305-20050}
Hunter Lightman, Vineet Kosaraju, Yura Burda, Harrison Edwards, Bowen Baker, Teddy Lee, Jan Leike, John Schulman, Ilya Sutskever, and Karl Cobbe. 2023.
\newblock \href {https://doi.org/10.48550/ARXIV.2305.20050} {Let's verify step by step}.
\newblock \emph{CoRR}, abs/2305.20050.

\bibitem[{Lin et~al.(2021)Lin, Wu, Yang, Lee, and Ren}]{DBLP:conf/acl/LinWYLR21}
Bill~Yuchen Lin, Ziyi Wu, Yichi Yang, Dong{-}Ho Lee, and Xiang Ren. 2021.
\newblock \href {https://doi.org/10.18653/V1/2021.FINDINGS-ACL.131} {Riddlesense: Reasoning about riddle questions featuring linguistic creativity and commonsense knowledge}.
\newblock In \emph{Findings of the Association for Computational Linguistics: {ACL/IJCNLP} 2021, Online Event, August 1-6, 2021}, volume {ACL/IJCNLP} 2021 of \emph{Findings of {ACL}}, pages 1504--1515. Association for Computational Linguistics.

\bibitem[{Madaan et~al.(2023)Madaan, Tandon, Gupta, Hallinan, Gao, Wiegreffe, Alon, Dziri, Prabhumoye, Yang, Welleck, Majumder, Gupta, Yazdanbakhsh, and Clark}]{DBLP:journals/corr/abs-2303-17651}
Aman Madaan, Niket Tandon, Prakhar Gupta, Skyler Hallinan, Luyu Gao, Sarah Wiegreffe, Uri Alon, Nouha Dziri, Shrimai Prabhumoye, Yiming Yang, Sean Welleck, Bodhisattwa~Prasad Majumder, Shashank Gupta, Amir Yazdanbakhsh, and Peter Clark. 2023.
\newblock \href {https://doi.org/10.48550/ARXIV.2303.17651} {Self-refine: Iterative refinement with self-feedback}.
\newblock \emph{CoRR}, abs/2303.17651.

\bibitem[{Mao et~al.(2024)Mao, Kim, and Zhou}]{DBLP:journals/corr/abs-2401-06961}
Yujun Mao, Yoon Kim, and Yilun Zhou. 2024.
\newblock \href {https://doi.org/10.48550/ARXIV.2401.06961} {{CHAMP:} {A} competition-level dataset for fine-grained analyses of llms' mathematical reasoning capabilities}.
\newblock \emph{CoRR}, abs/2401.06961.

\bibitem[{Nguyen et~al.(2024)Nguyen, Luo, Shiri, Phung, Li, Vu, and Haffari}]{DBLP:journals/corr/abs-2402-11199}
Minh{-}Vuong Nguyen, Linhao Luo, Fatemeh Shiri, Dinh Phung, Yuan{-}Fang Li, Thuy{-}Trang Vu, and Gholamreza Haffari. 2024.
\newblock \href {https://doi.org/10.48550/ARXIV.2402.11199} {Direct evaluation of chain-of-thought in multi-hop reasoning with knowledge graphs}.
\newblock \emph{CoRR}, abs/2402.11199.

\bibitem[{Oh et~al.(2024)Oh, Kim, Cha, and Oh}]{DBLP:journals/corr/abs-2402-06204}
Juhyun Oh, Eunsu Kim, Inha Cha, and Alice Oh. 2024.
\newblock \href {https://doi.org/10.48550/ARXIV.2402.06204} {The generative {AI} paradox on evaluation: What it can solve, it may not evaluate}.
\newblock \emph{CoRR}, abs/2402.06204.

\bibitem[{OpenAI(2023)}]{DBLP:journals/corr/abs-2303-08774}
OpenAI. 2023.
\newblock \href {https://doi.org/10.48550/ARXIV.2303.08774} {{GPT-4} technical report}.
\newblock \emph{CoRR}, abs/2303.08774.

\bibitem[{Pan et~al.(2023)Pan, Saxon, Xu, Nathani, Wang, and Wang}]{DBLP:journals/corr/abs-2308-03188}
Liangming Pan, Michael Saxon, Wenda Xu, Deepak Nathani, Xinyi Wang, and William~Yang Wang. 2023.
\newblock \href {https://doi.org/10.48550/ARXIV.2308.03188} {Automatically correcting large language models: Surveying the landscape of diverse self-correction strategies}.
\newblock \emph{CoRR}, abs/2308.03188.

\bibitem[{Paul et~al.(2023)Paul, Ismayilzada, Peyrard, Borges, Bosselut, West, and Faltings}]{DBLP:journals/corr/abs-2304-01904}
Debjit Paul, Mete Ismayilzada, Maxime Peyrard, Beatriz Borges, Antoine Bosselut, Robert West, and Boi Faltings. 2023.
\newblock \href {https://doi.org/10.48550/ARXIV.2304.01904} {{REFINER:} reasoning feedback on intermediate representations}.
\newblock \emph{CoRR}, abs/2304.01904.

\bibitem[{Saunders et~al.(2022)Saunders, Yeh, Wu, Bills, Ouyang, Ward, and Leike}]{DBLP:journals/corr/abs-2206-05802}
William Saunders, Catherine Yeh, Jeff Wu, Steven Bills, Long Ouyang, Jonathan Ward, and Jan Leike. 2022.
\newblock \href {https://doi.org/10.48550/ARXIV.2206.05802} {Self-critiquing models for assisting human evaluators}.
\newblock \emph{CoRR}, abs/2206.05802.

\bibitem[{Talmor et~al.(2018)Talmor, Herzig, Lourie, and Berant}]{DBLP:journals/corr/abs-1811-00937}
Alon Talmor, Jonathan Herzig, Nicholas Lourie, and Jonathan Berant. 2018.
\newblock \href {http://arxiv.org/abs/1811.00937} {Commonsenseqa: {A} question answering challenge targeting commonsense knowledge}.
\newblock \emph{CoRR}, abs/1811.00937.

\bibitem[{Team et~al.(2024)Team, Mesnard, Hardin, Dadashi, Bhupatiraju, Pathak, Sifre, Rivi{\`e}re, Kale, Love et~al.}]{team2024gemma}
Gemma Team, Thomas Mesnard, Cassidy Hardin, Robert Dadashi, Surya Bhupatiraju, Shreya Pathak, Laurent Sifre, Morgane Rivi{\`e}re, Mihir~Sanjay Kale, Juliette Love, et~al. 2024.
\newblock Gemma: Open models based on gemini research and technology.
\newblock \emph{arXiv preprint arXiv:2403.08295}.

\bibitem[{Touvron et~al.(2023)Touvron, Martin, Stone, Albert, Almahairi, Babaei, Bashlykov, Batra, Bhargava, Bhosale, Bikel, Blecher, Canton{-}Ferrer, Chen, Cucurull, Esiobu, Fernandes, Fu, Fu, Fuller, Gao, Goswami, Goyal, Hartshorn, Hosseini, Hou, Inan, Kardas, Kerkez, Khabsa, Kloumann, Korenev, Koura, Lachaux, Lavril, Lee, Liskovich, Lu, Mao, Martinet, Mihaylov, Mishra, Molybog, Nie, Poulton, Reizenstein, Rungta, Saladi, Schelten, Silva, Smith, Subramanian, Tan, Tang, Taylor, Williams, Kuan, Xu, Yan, Zarov, Zhang, Fan, Kambadur, Narang, Rodriguez, Stojnic, Edunov, and Scialom}]{DBLP:journals/corr/abs-2307-09288}
Hugo Touvron, Louis Martin, Kevin Stone, Peter Albert, Amjad Almahairi, Yasmine Babaei, Nikolay Bashlykov, Soumya Batra, Prajjwal Bhargava, Shruti Bhosale, Dan Bikel, Lukas Blecher, Cristian Canton{-}Ferrer, Moya Chen, Guillem Cucurull, David Esiobu, Jude Fernandes, Jeremy Fu, Wenyin Fu, Brian Fuller, Cynthia Gao, Vedanuj Goswami, Naman Goyal, Anthony Hartshorn, Saghar Hosseini, Rui Hou, Hakan Inan, Marcin Kardas, Viktor Kerkez, Madian Khabsa, Isabel Kloumann, Artem Korenev, Punit~Singh Koura, Marie{-}Anne Lachaux, Thibaut Lavril, Jenya Lee, Diana Liskovich, Yinghai Lu, Yuning Mao, Xavier Martinet, Todor Mihaylov, Pushkar Mishra, Igor Molybog, Yixin Nie, Andrew Poulton, Jeremy Reizenstein, Rashi Rungta, Kalyan Saladi, Alan Schelten, Ruan Silva, Eric~Michael Smith, Ranjan Subramanian, Xiaoqing~Ellen Tan, Binh Tang, Ross Taylor, Adina Williams, Jian~Xiang Kuan, Puxin Xu, Zheng Yan, Iliyan Zarov, Yuchen Zhang, Angela Fan, Melanie Kambadur, Sharan Narang, Aur{\'{e}}lien Rodriguez, Robert Stojnic, Sergey Edunov,
  and Thomas Scialom. 2023.
\newblock \href {https://doi.org/10.48550/ARXIV.2307.09288} {Llama 2: Open foundation and fine-tuned chat models}.
\newblock \emph{CoRR}, abs/2307.09288.

\bibitem[{Uesato et~al.(2022{\natexlab{a}})Uesato, Kushman, Kumar, Song, Siegel, Wang, Creswell, Irving, and Higgins}]{uesato2022solving}
Jonathan Uesato, Nate Kushman, Ramana Kumar, Francis Song, Noah Siegel, Lisa Wang, Antonia Creswell, Geoffrey Irving, and Irina Higgins. 2022{\natexlab{a}}.
\newblock Solving math word problems with process-and outcome-based feedback.
\newblock \emph{arXiv preprint arXiv:2211.14275}.

\bibitem[{Uesato et~al.(2022{\natexlab{b}})Uesato, Kushman, Kumar, Song, Siegel, Wang, Creswell, Irving, and Higgins}]{DBLP:journals/corr/abs-2211-14275}
Jonathan Uesato, Nate Kushman, Ramana Kumar, H.~Francis Song, Noah~Y. Siegel, Lisa Wang, Antonia Creswell, Geoffrey Irving, and Irina Higgins. 2022{\natexlab{b}}.
\newblock \href {https://doi.org/10.48550/ARXIV.2211.14275} {Solving math word problems with process- and outcome-based feedback}.
\newblock \emph{CoRR}, abs/2211.14275.

\bibitem[{Wang et~al.(2023)Wang, Li, Chen, Song, Lin, Cao, Liu, and Sui}]{DBLP:journals/corr/abs-2309-02144}
Peiyi Wang, Lei Li, Liang Chen, Feifan Song, Binghuai Lin, Yunbo Cao, Tianyu Liu, and Zhifang Sui. 2023.
\newblock \href {https://doi.org/10.48550/ARXIV.2309.02144} {Making large language models better reasoners with alignment}.
\newblock \emph{CoRR}, abs/2309.02144.

\bibitem[{Welleck et~al.(2023)Welleck, Lu, West, Brahman, Shen, Khashabi, and Choi}]{DBLP:conf/iclr/WelleckLWBSK023}
Sean Welleck, Ximing Lu, Peter West, Faeze Brahman, Tianxiao Shen, Daniel Khashabi, and Yejin Choi. 2023.
\newblock \href {https://openreview.net/pdf?id=hH36JeQZDaO} {Generating sequences by learning to self-correct}.
\newblock In \emph{The Eleventh International Conference on Learning Representations, {ICLR} 2023, Kigali, Rwanda, May 1-5, 2023}. OpenReview.net.

\bibitem[{Weng et~al.(2023)Weng, Zhu, Xia, Li, He, Liu, Sun, Liu, and Zhao}]{DBLP:conf/emnlp/WengZX0HLSLZ23}
Yixuan Weng, Minjun Zhu, Fei Xia, Bin Li, Shizhu He, Shengping Liu, Bin Sun, Kang Liu, and Jun Zhao. 2023.
\newblock \href {https://aclanthology.org/2023.findings-emnlp.167} {Large language models are better reasoners with self-verification}.
\newblock In \emph{Findings of the Association for Computational Linguistics: {EMNLP} 2023, Singapore, December 6-10, 2023}, pages 2550--2575. Association for Computational Linguistics.

\bibitem[{West et~al.(2023)West, Lu, Dziri, Brahman, Li, Hwang, Jiang, Fisher, Ravichander, Chandu, Newman, Koh, Ettinger, and Choi}]{DBLP:journals/corr/abs-2311-00059}
Peter West, Ximing Lu, Nouha Dziri, Faeze Brahman, Linjie Li, Jena~D. Hwang, Liwei Jiang, Jillian Fisher, Abhilasha Ravichander, Khyathi Chandu, Benjamin Newman, Pang~Wei Koh, Allyson Ettinger, and Yejin Choi. 2023.
\newblock \href {https://doi.org/10.48550/ARXIV.2311.00059} {The generative {AI} paradox: "what it can create, it may not understand"}.
\newblock \emph{CoRR}, abs/2311.00059.

\bibitem[{Wu et~al.(2023)Wu, Hu, Shi, Dziri, Suhr, Ammanabrolu, Smith, Ostendorf, and Hajishirzi}]{DBLP:journals/corr/abs-2306-01693}
Zeqiu Wu, Yushi Hu, Weijia Shi, Nouha Dziri, Alane Suhr, Prithviraj Ammanabrolu, Noah~A. Smith, Mari Ostendorf, and Hannaneh Hajishirzi. 2023.
\newblock \href {https://doi.org/10.48550/ARXIV.2306.01693} {Fine-grained human feedback gives better rewards for language model training}.
\newblock \emph{CoRR}, abs/2306.01693.

\bibitem[{Xu et~al.(2024)Xu, Diao, Yang, and Wang}]{DBLP:journals/corr/abs-2402-10528}
Xin Xu, Shizhe Diao, Can Yang, and Yang Wang. 2024.
\newblock \href {https://doi.org/10.48550/ARXIV.2402.10528} {Can we verify step by step for incorrect answer detection?}
\newblock \emph{CoRR}, abs/2402.10528.

\bibitem[{Yang et~al.(2024)Yang, Gribovskaya, Kassner, Geva, and Riedel}]{DBLP:journals/corr/abs-2402-16837}
Sohee Yang, Elena Gribovskaya, Nora Kassner, Mor Geva, and Sebastian Riedel. 2024.
\newblock \href {https://doi.org/10.48550/ARXIV.2402.16837} {Do large language models latently perform multi-hop reasoning?}
\newblock \emph{CoRR}, abs/2402.16837.

\bibitem[{Ye et~al.(2023)Ye, Jo, Kim, Kim, Hwang, and Seo}]{selfee2023}
Seonghyeon Ye, Yongrae Jo, Doyoung Kim, Sungdong Kim, Hyeonbin Hwang, and Minjoon Seo. 2023.
\newblock \href {https://kaistai.github.io/SelFee/} {Selfee: Iterative self-revising llm empowered by self-feedback generation}.
\newblock Blog post.

\bibitem[{Yu et~al.(2023{\natexlab{a}})Yu, Gao, and Wang}]{DBLP:journals/corr/abs-2311-09724}
Fei Yu, Anningzhe Gao, and Benyou Wang. 2023{\natexlab{a}}.
\newblock \href {https://doi.org/10.48550/ARXIV.2311.09724} {Outcome-supervised verifiers for planning in mathematical reasoning}.
\newblock \emph{CoRR}, abs/2311.09724.

\bibitem[{Yu et~al.(2023{\natexlab{b}})Yu, Peng, Galley, Gao, and Yu}]{DBLP:journals/corr/abs-2310-13522}
Xiao Yu, Baolin Peng, Michel Galley, Jianfeng Gao, and Zhou Yu. 2023{\natexlab{b}}.
\newblock \href {https://doi.org/10.48550/ARXIV.2310.13522} {Teaching language models to self-improve through interactive demonstrations}.
\newblock \emph{CoRR}, abs/2310.13522.

\bibitem[{Yuan et~al.(2024)Yuan, Pang, Cho, Sukhbaatar, Xu, and Weston}]{DBLP:journals/corr/abs-2401-10020}
Weizhe Yuan, Richard~Yuanzhe Pang, Kyunghyun Cho, Sainbayar Sukhbaatar, Jing Xu, and Jason Weston. 2024.
\newblock \href {https://doi.org/10.48550/ARXIV.2401.10020} {Self-rewarding language models}.
\newblock \emph{CoRR}, abs/2401.10020.

\bibitem[{Yuan et~al.(2023)Yuan, Yuan, Li, Dong, Tan, and Zhou}]{DBLP:journals/corr/abs-2308-01825}
Zheng Yuan, Hongyi Yuan, Chengpeng Li, Guanting Dong, Chuanqi Tan, and Chang Zhou. 2023.
\newblock \href {https://doi.org/10.48550/ARXIV.2308.01825} {Scaling relationship on learning mathematical reasoning with large language models}.
\newblock \emph{CoRR}, abs/2308.01825.

\bibitem[{Y{\"{u}}ksekg{\"{o}}n{\"{u}}l et~al.(2023)Y{\"{u}}ksekg{\"{o}}n{\"{u}}l, Chandrasekaran, Jones, Gunasekar, Naik, Palangi, Kamar, and Nushi}]{DBLP:journals/corr/abs-2309-15098}
Mert Y{\"{u}}ksekg{\"{o}}n{\"{u}}l, Varun Chandrasekaran, Erik Jones, Suriya Gunasekar, Ranjita Naik, Hamid Palangi, Ece Kamar, and Besmira Nushi. 2023.
\newblock \href {https://doi.org/10.48550/ARXIV.2309.15098} {Attention satisfies: {A} constraint-satisfaction lens on factual errors of language models}.
\newblock \emph{CoRR}, abs/2309.15098.

\bibitem[{Zelikman et~al.(2022)Zelikman, Wu, Mu, and Goodman}]{DBLP:conf/nips/ZelikmanWMG22}
Eric Zelikman, Yuhuai Wu, Jesse Mu, and Noah~D. Goodman. 2022.
\newblock \href {http://papers.nips.cc/paper\_files/paper/2022/hash/639a9a172c044fbb64175b5fad42e9a5-Abstract-Conference.html} {Star: Bootstrapping reasoning with reasoning}.
\newblock In \emph{Advances in Neural Information Processing Systems 35: Annual Conference on Neural Information Processing Systems 2022, NeurIPS 2022, New Orleans, LA, USA, November 28 - December 9, 2022}.

\bibitem[{Zhang et~al.(2024)Zhang, Blum, Choji, Shah, and Vempala}]{Zhang2024ULTRAUL}
Xinliang~Frederick Zhang, Carter Blum, Temma Choji, Shalin Shah, and Alakananda Vempala. 2024.
\newblock \href {https://api.semanticscholar.org/CorpusID:267199878} {Ultra: Unleash llms' potential for event argument extraction through hierarchical modeling and pair-wise refinement}.
\newblock \emph{ArXiv}, abs/2401.13218.

\bibitem[{Zheng et~al.(2024{\natexlab{a}})Zheng, Yin, Zhou, Meng, Zhou, Chang, Huang, and Peng}]{zheng2024prompt}
Chujie Zheng, Fan Yin, Hao Zhou, Fandong Meng, Jie Zhou, Kai-Wei Chang, Minlie Huang, and Nanyun Peng. 2024{\natexlab{a}}.
\newblock Prompt-driven llm safeguarding via directed representation optimization.
\newblock \emph{arXiv preprint arXiv:2401.18018}.

\bibitem[{Zheng et~al.(2024{\natexlab{b}})Zheng, Zhang, Zhang, Ye, Luo, and Ma}]{llama-factory}
Yaowei Zheng, Richong Zhang, Junhao Zhang, Yanhan Ye, Zheyan Luo, and Yongqiang Ma. 2024{\natexlab{b}}.
\newblock \href {http://arxiv.org/abs/2403.13372} {Llamafactory: Unified efficient fine-tuning of 100+ language models}.
\newblock \emph{arXiv preprint arXiv:2403.13372}.

\bibitem[{Zhou et~al.(2023)Zhou, Wang, Lu, Shi, Luo, Qin, Lu, Jia, Song, Zhan, and Li}]{DBLP:journals/corr/abs-2308-07921}
Aojun Zhou, Ke~Wang, Zimu Lu, Weikang Shi, Sichun Luo, Zipeng Qin, Shaoqing Lu, Anya Jia, Linqi Song, Mingjie Zhan, and Hongsheng Li. 2023.
\newblock \href {https://doi.org/10.48550/ARXIV.2308.07921} {Solving challenging math word problems using {GPT-4} code interpreter with code-based self-verification}.
\newblock \emph{CoRR}, abs/2308.07921.

\bibitem[{Zou et~al.(2023)Zou, Phan, Chen, Campbell, Guo, Ren, Pan, Yin, Mazeika, Dombrowski, Goel, Li, Byun, Wang, Mallen, Basart, Koyejo, Song, Fredrikson, Kolter, and Hendrycks}]{DBLP:journals/corr/abs-2310-01405}
Andy Zou, Long Phan, Sarah Chen, James Campbell, Phillip Guo, Richard Ren, Alexander Pan, Xuwang Yin, Mantas Mazeika, Ann{-}Kathrin Dombrowski, Shashwat Goel, Nathaniel Li, Michael~J. Byun, Zifan Wang, Alex Mallen, Steven Basart, Sanmi Koyejo, Dawn Song, Matt Fredrikson, J.~Zico Kolter, and Dan Hendrycks. 2023.
\newblock \href {https://doi.org/10.48550/ARXIV.2310.01405} {Representation engineering: {A} top-down approach to {AI} transparency}.
\newblock \emph{CoRR}, abs/2310.01405.

\end{thebibliography}

\appendix
\onecolumn
\section{Prompts}\label{sec:prompts}

\begin{tcolorbox}[title=8-Shot CoT Prompts with 2 Shots Shown Below]
\small
Q: Jason had 20 lollipops. He gave Denny some lollipops. Now Jason has 12 lollipops. How many lollipops did Jason give to Denny?

Explain your reasoning step by step. Your final answer should be a single numerical number, in the form \boxed{answer}. End your response with [END].
\\

Step 1: Jason originally had 20 lollipops.

Step 2: After giving some to Denny, Jason has 12 lollipops.

Step 3: The number of lollipops Jason gave to Denny is the difference between the original number and the remaining number, which is 20 - 12 = 8 lollipops.

Step 4: The answer is \boxed{8} lollipops. [END]
\\

Q: There are 15 trees in the grove. Grove workers will plant trees in the grove today. After they are done, there will be 21 trees. How many trees did the grove workers plant today?

Explain your reasoning step by step. Your final answer should be a single numerical number, in the form \boxed{answer}. End your response with [END].
\\

Step 1: Initially, there are 15 trees in the grove.

Step 2: After planting, the total number of trees in the grove is 21.

Step 3: The number of trees planted by the grove workers today is the difference between the final number and the initial number of trees, which is 21 - 15 = 6 trees.

Step 4: The answer is \boxed{6} trees. [END]
\end{tcolorbox}

\begin{tcolorbox}[title=2-Shot Critique Prompts with 1 Shot Shown Below]
\small
Q: Betty is saving money for a new wallet which costs \$100. Betty has only half of the money she needs. Her parents decided to give her \$15 for that purpose, and her grandparents twice as much as her parents. How much more money does Betty need to buy the wallet?
\\

Answer 1 (Incorrect):
\\

Step 1: Betty needs \$100 to buy the wallet.

Step 2: Her parents gave her \$15, which is half of the remaining amount.

...

Step 7: The answer is \boxed{\$55}.
\\

Answer 2 (Correct):
\\

Step 1: Betty needs \$100 to buy the wallet, but she currently has only half of the money, which is \$50.

Step 2: Her parents gave her \$15, so she now has \$50 + \$15 = \$65.

...

Step 6: The answer is \boxed{5} dollars.
\\

There are reasoning errors in Answer 1. Please go through each
step in Answer 1, use Answer 2 as a reference for the correct
approach, and provide feedback that helps correct the errors in
Answer 1. End your response with [END].
\\

Let's go through the errors in Answer 1 and provide feedback:
\\

Answer 1 (Incorrect):
\\

Step 1: Betty needs \$100 to buy the wallet.

Feedback: This step is correct.
\\

Step 2: Her parents gave her \$15, which is half of the remaining amount.

Feedback: This is incorrect. Betty's parents gave her \$15, but it's not necessarily half of the remaining amount. The question states that Betty has only half of the money she needs, not half of the remaining amount after her parents' contribution.
\\

...
\\

Step 7: The answer is \boxed{\$55}.

Feedback: The correct answer, based on the corrected calculations, should be \boxed{\$5}. [END]
\end{tcolorbox}

\begin{tcolorbox}[title=2-Shot Correction Prompts with 1 Shot Shwon Below]
\small
Q: Kelly is grocery shopping at a supermarket and is making sure she has enough in her budget for the items in her cart. Her 5 packs of bacon cost \$10 in total and she has 6 packets of chicken which each cost twice as much as a pack of bacon. She also has 3 packs of strawberries, priced at \$4 each, and 7 packs of apples, each priced at half the price of a pack of strawberries. If Kelly’s budget is \$65 then how much money, in dollars, does she have left in her budget?
\\

Step 1: Kelly spends \$10 in total for 5 packs of bacon.

Feedback: This step is correct.
\\

Step 2: Each packet of chicken costs twice as much as a pack of bacon, which is 2 x \$10 = \$20.

Feedback: Each pack of chicken costs twice as much as a pack of bacon, not the total cost of bacon. So we need to calculate the cost of one pack of bacon first (total cost of bacon / total bags of bacon) and use that.
\\

...
\\

Step 7: The answer is \boxed{\$29}.
\\

Can you correct the errors in your reasoning based on the feedback provided? Your final answer should be a single numerical number, in the form \boxed{answer}. End your response with [END].
\\

Step 1: Kelly spends \$10 in total for 5 packs of bacon, so each pack of bacon costs \$10 / 5 = \$2.

Step 2: Each packet of chicken costs twice as much as a pack of bacon, which is 2 x \$2 = \$4. Since she has 6 packets of
chicken, she spends 6 x \$4 = \$24 on chicken.

...

Step 7: The answer is \boxed{\$5}. [END]
\end{tcolorbox}

\begin{tcolorbox}[title=4-Shot Verifying Prompts with 2 Shots Shown Below]
\small
Question: Betty is saving money for a new wallet which costs \$100. Betty has only half of the money she needs. Her parents
decided to give her \$15 for that purpose, and her grandparents twice as much as her parents. How much more money does
Betty need to buy the wallet?
\\

Solution:

Step 1: Betty needs \$100 to buy the wallet.

Step 2: Her parents gave her \$15, which is half of the remaining amount.

Step 3: Her grandparents gave her twice as much as her parents, which is 2 x \$15 = \$30.

Step 4: The total amount of money Betty has now is \$15 + \$30 = \$45.

Step 5: The difference between the amount she needs and the amount she has is \$100 - \$45 = \$55.

Step 6: Betty still needs \$55 to buy the wallet.

Step 7: The answer is \boxed{\$55}.
\\

Is this solution correct or not? Answer **correct** or **incorrect** and explain your reasoning. End your response with
[END].
\\

The solution is **incorrect**. You've ignored the original amount that Betty had saved, which is half the wallet's cost (\$100/2
= \$50). So, before receiving money from her parents and grandparents, she already had \$50. [END]
\\

Question: Betty is saving money for a new wallet which costs \$100. Betty has only half of the money she needs. Her parents
decided to give her \$15 for that purpose, and her grandparents twice as much as her parents. How much more money does
Betty need to buy the wallet?
\\

Solution:

Step 1: Betty needs \$100 to buy the wallet, but she currently has only half of the money, which is \$50.

Step 2: Her parents gave her \$15, so she now has \$50 + \$15 = \$65.

Step 3: Her grandparents gave her twice as much as her parents, so she received \$15 x 2 = \$30 from her grandparents.

Step 4: Betty now has a total of \$65 + \$30 = \$95.

Step 5: To find out how much more money Betty needs, subtract the amount she has from the amount she needs, which is
\$100 - \$95 = \$5.

Step 6: The answer is \boxed{5} dollars.
\\

Is this solution correct or not? Answer **correct** or **incorrect** and explain your reasoning. End your response with
[END].
\\

This solution is **correct**. Betty only needs \$5 more to buy the \$100 wallet. This is because she started with \$50, received
\$15 from her parents, and received \$30 from her grandparents, which totals \$95. Subtracting that from the total cost of the
wallet leaves her with needing just \$5. [END]

\end{tcolorbox}

\twocolumn
\section{Proof: Using Correct Solutions as Hints Make Critique Generation Easier}\label{app:proof}
Intuitively, if we provide the LM with a correct solution, it will be easier for the LM to generate a critique for the incorrect solution of the same question.
In fact, we can verify this intuition from a mathematical perspective. 
Given a pair of incorrect-correct solutions $(s^{-}, s^{+})$, our goal is to learn a mapping from the incorrect solution to the correct one, which is modeled by $M(s^{+}|q,s^{-})$, where $M(\cdot)$ is the probability distribution of the base LM.
By introducing critique $c$ as intermediate generations, we have:
\begin{equation}
    M(s^{+}|q,s^{-}) = M(c|q,s^{-}) \cdot M(s^{+}|q,s^{-},c)
\end{equation}
The optimal critique $c^{*}$ that we want to find should be best at facilitating this two-phase generation process, i.e.,
\begin{equation}
    c^{*}=\argmax_c M(c|q,s^{-}) \cdot M(s^{+}|q,s^{-},c)
\end{equation}
To obtain the optimal critique, we can first ask the model to reverse-engineer a critique $\hat{c}$ with the correct solution $s^{+}$ as a hint:
\begin{equation}\label{eq:critique_generation}
    \hat{c}=\argmax_c M(c|q,s^{-},s^{+})
\end{equation}
With Bayes' theorem, Equation~\ref{eq:critique_generation} can be re-written as:
\begin{equation}
\begin{aligned}
     \hat{c} & =\argmax_c M(c|q,s^{-},s^{+}) \\
     & =\argmax_c \frac{M(c|q,s^{-}) \cdot M(s^{+}|q,s^{-},c)}{M(s^{+}|q,s^{-})} \\
     & =\argmax_c M(c|q,s^{-}) \cdot M(s^{+}|q,s^{-},c) \\
     & = c^{*}
\end{aligned}
\end{equation}
which is exactly what we want. 

This simple proof shows that in principle, the critique generated with the correct solution as a hint should be best at guiding the LM to recover the correct solution from the incorrect one.

\section{Combining \model with Rejection Sampling Fine-Tuning}\label{sec:rft_detail}
Aside from self-correction upon the initial solution generated by the base LM as shown in Figure~\ref{fig:new_main_results}, we also explore whether our proposed self-correction method can be combined with other fine-tuning methods (e.g., rejection sampling fine-tuning) to further improve reasoning performance. Here we replace the base LM with the RFT model.
Rejection sampling fine-tuning leverages correct generations for training. Yet it ignores rich information in the large amounts of incorrect solutions. We hope weaker LMs can also learn from its own mistakes to become better reasoners. 
Since we have already obtained a stronger RFT model that avoids some mistakes by base LMs that are easier to fix, we want our refiner to learn to correct errors that require more in-depth thinking. Consequently, to collect the incorrect solutions to be reflected upon, we \textit{resample} 10 solutions
for each question from the RFT model, instead of reusing the sampled solutions from the base LM in step (a) of Figure~\ref{fig:method}. Then we follow the rest of the pipeline to collect critique-correction data for refiner finetuning.

\begin{table}\setlength\tabcolsep{3pt}\centering
\small
\begin{tabular}{lc|cccc}\toprule
\multirow{2}{*}{Verifier} &\multirow{2}{*}{Refiner} &\multicolumn{2}{c}{GSM8K} \\\cmidrule{3-4}
 & &Verifier F1 &Final Accu. \\\midrule
  \multicolumn{2}{l}{Initial solutions by RFT model} &N/A &42.7 \\\cmidrule{1-2}
ours &\multirow{5}{*}{\model} &47.3 &42.8 \hlc[fullgreen!15]{(+0.1)} \\
\texttt{gpt-3.5-turbo} & &59.7 &31.5 \hlc[fullred!90]{(-11.2)} \\
\quad w/ SC@10 & &63.6 &38.1 \hlc[fullred!60]{(-4.6)} \\
\texttt{gpt-4} & &\textbf{89.2} &\textbf{46.3} \hlc[fullgreen!40]{(+3.6)} \\
oracle & &100.0 &51.4 \hlc[fullgreen!70]{(+8.7)}\\
\bottomrule
\end{tabular}
\caption{Performance of self-correction with different inputs on GSM8K test set using \texttt{LLaMA-2-13B-chat} as base LM. The initial solution is generated by the rejection-sampling fine-tuned model.}\label{tab:main_results}
\end{table}

\begin{table}\setlength\tabcolsep{3pt}\centering
\small
\begin{tabular}{ccccc}\toprule
\makecell[c]{Verifier\\param.\\init.\\from} &\makecell[c]{Refiner\\param.\\init.\\from} &\makecell[c]{Refiner\\ft. on\\solutions\\from} &\makecell[c]{Final\\Accu.} &\makecell[c]{Oracle\\Accu.} \\\cmidrule{1-5}
\multicolumn{3}{c}{Initial solutions by RFT model} &44.9 &44.9 \\\cmidrule{1-5}
RFT & base &RFT &\textbf{45.3} &\textbf{53.3} \\
base &base &RFT &45.0 &\textbf{53.3}\\
RFT &RFT &RFT &45.2 &52.3 \\
base &RFT &RFT &45.0 &52.3 \\
RFT &base &base &45.1 &\textbf{53.3} \\
\bottomrule
\end{tabular}
\caption{Ablation study on parameter initialization and refiner's fine-tuning data source. Results are shown on GSM8K dev set with \texttt{LLaMA-2-13B-chat} as base LM. 
}\label{tab:ablation_rft}
\end{table}

\begin{table*}[h]\setlength\tabcolsep{2pt}\centering
\footnotesize
\begin{tabular}{lccccccccccccc}\toprule
\multirow{2}{*}{Verifier} &\multirow{2}{*}{Refiner} &\multicolumn{2}{c}{GSM8K} &\multicolumn{2}{c}{\makecell[c]{GSM8K $\rightarrow$ \\MATH Subset}} &\multicolumn{2}{c}{CSQA} &\multicolumn{2}{c}{\makecell[c]{CSQA $\rightarrow$ \\QASC}} &\multicolumn{2}{c}{\makecell[c]{CSQA $\rightarrow$ \\RiddleSense}}  \\\cmidrule{3-12}
 & &V. F1 &ACC &V. F1 &ACC &V. F1 &ACC &V. F1 &ACC &V. F1 &ACC \\\midrule
 \multicolumn{4}{l}{\textit{Base LM: \texttt{LLaMA-2-13B-chat}}} \\\cmidrule{1-12}
\multicolumn{2}{l}{\makecell[l]{Initial answers by\\few-shot prompting}} &- &37.2 &-  & 23.8 &- & 69.7 & - & 65.9 &-  & 57.6\\\cmidrule{1-2}
prompted &\multirow{2}{*}{prompted} &31.9 &36.9 $_{\text{-0.3}}$ & 20.0 & 23.8 $_{\text{+0.0}}$ & 52.5 & 62.2 $_{\text{-7.5}}$ & 53.1 & 63.3 $_{\text{-2.6}}$ & 51.9 & 55.2 $_{\text{-2.4}}$ \\
oracle & &100.0 &39.7 $_{\text{+2.5}}$ & 100.0 & 26.5 $_{\text{+2.7}}$ & 100.0 & 83.7 $_{\text{+14.0}}$ & 100.0 & 76.1 $_{\text{+10.2}}$ & 100.0 & 72.7 $_{\text{+15.1}}$\\\cmidrule{1-2}
self &\multirow{3}{*}{\makecell[c]{\model\\(fine-tuned)}} &51.1 &37.5 $_{\text{+0.3}}$ & 36.0 & 23.8 $_{\text{+0.0}}$  & 47.1 &69.7 $_{\text{+0.0}}$ & 45.8 & 65.6 $_{\text{-0.3}}$ & 42.1 & 57.5 $_{\text{-0.1}}$ \\
\cellcolor{maroon!10} \texttt{gpt-4} & & \cellcolor{maroon!10}88.4 &\cellcolor{maroon!10}41.4 $_{\text{+4.2}}$ &\cellcolor{maroon!10} 82.9 &\cellcolor{maroon!10} 26.5 $_{\text{+2.7}}$  &\cellcolor{maroon!10} 80.3 &\cellcolor{maroon!10} 72.4 $_{\text{+2.7}}$ & \cellcolor{maroon!10}85.9 &\cellcolor{maroon!10} 68.1 $_{\text{+2.2}}$ &\cellcolor{maroon!10} 87.7 &\cellcolor{maroon!10} \textbf{67.3} $_{\text{+9.7}}$ \\
oracle & &100.0 &46.0 $_{\text{+8.8}}$ & 100.0 & 31.5 $_{\text{+7.7}}$  & 100.0 & 86.2 $_{\text{+16.5}}$ & 100.0 & 78.0 $_{\text{+12.1}}$ & 100.0 & 76.4 $_{\text{+18.8}}$ \\
\midrule
 \multicolumn{4}{l}{\textit{Base LM: \texttt{Gemma-7B-it}}} \\\cmidrule{1-12}
\multicolumn{2}{l}{\makecell[l]{Initial answers by\\few-shot prompting}} &- &36.3 &- & 27.1  &- & 67.2 &- & 65.0 &-  & 57.1\\\cmidrule{1-2}
prompted &\multirow{2}{*}{prompted} &45.3 &36.3 $_{\text{+0.0}}$ & 39.2 & 28.2 $_{\text{+1.1}}$  &55.7 & 65.4 $_{\text{-1.8}}$ & 61.0 & 65.1 $_{\text{+0.1}}$ & 58.2 & 55.3 $_{\text{-1.8}}$ \\
oracle & &100.0 &37.7 $_{\text{+1.4}}$ & 100.0 & 29.3 $_{\text{+2.2}}$  & 100.0 & 74.5 $_{\text{+7.3}}$ & 100.0 & 71.3 $_{\text{+6.3}}$ & 100.0 & 62.3 $_{\text{+5.2}}$ \\\cmidrule{1-2}
self &\multirow{3}{*}{\makecell[c]{\model\\(fine-tuned)}} &56.8 &36.7 $_{\text{+0.4}}$ & 49.5 & 27.1 $_{\text{+0.0}}$  & 42.0 &67.5 $_{\text{+0.3}}$ & 41.0 & 65.2 $_{\text{+0.2}}$ & 38.1 & 56.8 $_{\text{-0.3}}$ \\
\cellcolor{maroon!10}\texttt{gpt-4} & &\cellcolor{maroon!10} 89.5 &\cellcolor{maroon!10} \textbf{42.5} $_{\text{+6.2}}$ &\cellcolor{maroon!10} 82.8 &\cellcolor{maroon!10} \textbf{39.2} $_{\text{+12.1}}$  &\cellcolor{maroon!10} 82.7 &\cellcolor{maroon!10} \textbf{75.0} $_{\text{+7.8}}$ &\cellcolor{maroon!10} 89.9 &\cellcolor{maroon!10} \textbf{72.8} $_{\text{+7.8}}$ &\cellcolor{maroon!10} 85.6 &\cellcolor{maroon!10} 64.9 $_{\text{+7.8}}$  \\
oracle & &100.0 &47.4 $_{\text{+11.1}}$ & 100.0 & 44.2 $_{\text{+17.1}}$  & 100.0 & 85.4 $_{\text{+18.2}}$ & 100.0 & 77.3 $_{\text{+12.3}}$ & 100.0 & 72.7 $_{\text{+15.6}}$ \\
\bottomrule
\end{tabular}
\caption{
Performance of \model models using \texttt{LLaMA-2-13B-chat} and \texttt{Gemma-7B-it} as base LM.
We report F1 score of the verifiers (V. F1) and  final answer accuracy (ACC).
We include test results for training tasks (GSM8K and CommonsenseQA/CSQA), as well as transfer evaluation of GSM8K trained models on MATH subset, CSQA trained models on QASC and RiddleSense. 
All models use greedy decoding. 
We \hlc[maroon!10]{highlight} the best-performing system per model without using an oracle verifier.
On each dataset, the superior model among the \hlc[maroon!10]{highlighted} ones is indicated in \textbf{bold}.
}\label{tab:new_main_results}
\end{table*}

Table~\ref{tab:main_results} shows that our method is complementary to RFT. Concretely, RFT improves the final accuracy upon the few-shot prompting baseline from 37.2 to 42.7, while our self-correction system (gpt-4 as verifier and our finetuned LLaMA as refiner) can further improve the performance from 42.7 to 45.1, demonstrating the effectiveness of our method over the few-shot prompting and RFT baselines.

Table~\ref{tab:ablation_rft} investigates the optimal approach for initializing the parameters of the refiner and verifier during fine-tuning. Our empirical findings indicate that initializing the verifier from the RFT model and the refiner from the base language model results in superior performance. We hypothesize that stronger reasoning capabilities of RFT model could complement the verification skills.  Furthermore, when generating critiques of solutions, those created using solutions by the RFT model outperform those based on the base model.

\section{Combining \model with Oversample-then-Rerank}\label{sec:comp_rerank}


Self-correction could be combined with oversample-then-rerank, an orthogonal approach to further improve reasoning~\cite{DBLP:journals/corr/abs-2110-14168,DBLP:conf/acl/LiLZFCLC23,Hosseini2024VSTaRTV}. The oversample-then-rerank approach first samples $k$ solutions per question by few-shot prompting, then leverages a trained verifier to score the correctness of each solution, and selects the final answer based on a weighted voting scheme~\cite{DBLP:conf/acl/LiLZFCLC23} as follows:
\begin{align*}
\hat{\mathbf{y}}=\underset{\mathbf{y}}{\arg\max}\sum_{i=1}^k{\mathbbm{1}_{\mathbf{y}_i=\mathbf{y}}\cdot p_v(q,s_i,\mathbf{y}_i)},
\end{align*}
where $q$ is the question, $s_i$ is the $i$-th sampled solutions, $\mathbf{y}_i$ is the final answer extracted from $s_i$, and $\mathbbm{1}_{\mathbf{y}_i=\mathbf{y}}$ is an indicator function that returns 1 (or 0) if $\mathbf{y}_i=\mathbf{y}$ (or not), and $p_v(\cdot)$ is the probability produced by the verifier\footnote{We find that for answer aggregation, weighted voting performs better than majority voting and selecting the top-1 solution, which echoes the findings of ~\citet{DBLP:conf/acl/LiLZFCLC23}. Therefore, we report the weighted voting results in this section.}

To integrate self-correction with the oversample-then-rerank approach, we apply the \model refiner to solutions predicted to be correct with a probability \textit{below} a certain threshold. This threshold is automatically determined to maximize accuracy on the development set and is then fixed for use during test-time predictions. The refiner greedily decodes one refined solution for each solution. Subsequently, we combine the refined solutions with the original solutions predicted as correct to arrive at the final answer using weighted voting for aggregation:

\begin{align*}
\hat{\mathbf{y}}=\underset{\mathbf{y}}{\arg\max}\sum_{i=1}^k\bigl(&\mathbbm{1}_{\mathbf{y}_i=\mathbf{y}}\cdot p_v^i \cdot \mathbbm{1}_{p_v^i \geq C} \\
&+ \mathbbm{1}_{{\mathbf{y}'}_i=\mathbf{y}}\cdot p_v^{i'} \cdot \mathbbm{1}_{p_v^{i}<C}\bigr),
\end{align*}
where $p_v^i=p_v(q,s_i,\mathbf{y}_i)$ is the verifier probability for the original solution $s_i$, $p_v^{i'}=p_v(q,s'_i,\mathbf{y'}_i)$ is the verifier probability for the refined solution $s'_i$ coming from $s_i$, and $C$ is a threshold for determining when to self-refine.

We sample $k=10$ solutions for each question with a temperature of 0.9 and use the same fine-tuned self-verifier (not \texttt{gpt-4} or oracle verifier) across all the settings in Table~\ref{tab:baseline_results} that involve a verifier. As shown by previous work, oversample-then-rerank improves performance upon few-shot prompting with only greedy decoding by a large margin, because it better explores the solution space by sampling multiple solutions.
Our self-correction method is both \textit{orthogonal} to and \textit{synergistic} with the oversample-then-rerank.
By incorporating our self-correction method into the baseline, with only a weak self-verifier, we still observe an improvement in accuracy compared to the baseline: +1.7 for \texttt{LLaMA-2-13B-chat} on GSM8K, and +1.6 for GSM8K-trained \texttt{Gemma-7B-it} on MATH. 
By aggregating verifications from multiple solutions, the weak verifiers become stronger via ensembling and thus unleash the power of refiner for better self-correction of reasoning.


\begin{table}\setlength\tabcolsep{3pt}\centering
\small
\begin{tabular}{cc|cc}\toprule
\makecell[c]{\# Candidate\\Solutions} &\makecell[c]{Refine w/\\SCORE?}  &GSM8K &\makecell[c]{GSM8K$\rightarrow$\\MATH Subset}  \\\midrule
\multicolumn{4}{l}{\textit{Base LM: \texttt{LLaMA-2-13B-chat}}} \\\cmidrule{1-4}
1 & \XSolidBrush & 37.2 &23.8  \\
1 & \Checkmark & 37.5 $_{\text{+0.3}}$ &23.8 $_{\text{+0.0}}$  \\
10 & \XSolidBrush & 44.3 $_{\text{+7.1}}$ &\textbf{30.9 $_{\text{+7.1}}$}  \\
10 & \Checkmark & \textbf{46.0 $_{\text{+8.8}}$} &29.8 $_{\text{+6.0}}$  \\\midrule
\multicolumn{4}{l}{\textit{Base LM: \texttt{Gemma-7B-it}}} \\\cmidrule{1-4}
1 & \XSolidBrush & 36.3 &27.1  \\
1 & \Checkmark & 36.7 $_{\text{+0.4}}$ &27.1 $_{\text{+0.0}}$  \\
10 & \XSolidBrush & 39.5 $_{\text{+3.2}}$ &34.3 $_{\text{+7.2}}$  \\
10 & \Checkmark & \textbf{41.1 $_{\text{+4.8}}$} &\textbf{35.9 $_{\text{+8.8}}$}  \\
\bottomrule
\end{tabular}
\caption{Self-correction combined with the oversample-then-rerank strategy~\cite{DBLP:journals/corr/abs-2110-14168} yield better results than using either one alone. We show results on GSM8K test set and transfer evaluation of GSM8K trained models on MATH subset using \texttt{LLaMA-2-13B-chat} and \texttt{Gemma-7B-it} as base LMs.}\label{tab:baseline_results}
\end{table}

\section{Inference Overhead by \model}\label{sec:latency}
Although we employ three models for inference (few-shot prompted solution generator, self-verifier, self-refiner), it will not necessarily triple the latency. The self-verifier's latency is minimal because it classifies rather than generates, and the refiner is invoked only if the verifier flags the initial solution as incorrect. The increased inference time of our pipeline over the base model is modest, at x1.3 times for \texttt{LLaMA-2-13B-chat} and x1.4 times for \texttt{Gemma-7B-it}. Comparatively, the oversample-then-rerank baseline (Appendix~\ref{sec:comp_rerank}) results in a 4.5 times increase in latency when sampling 10 solutions per question.

\end{document}